\documentclass[journal]{IEEEtran}
\usepackage{cite}

\usepackage{xcolor}
\usepackage{tabularx}
\usepackage{multirow}
\usepackage{graphicx}
\usepackage{dsfont}
\usepackage[normalem]{ulem}
\useunder{\uline}{\ul}{}
\usepackage{amsmath}
\usepackage{hyperref}

\begin{document}
%
\title{
Rethinking Affect Analysis: A Protocol for Ensuring Fairness and Consistency
}
%
%
%

\author{Guanyu~Hu, Dimitrios~Kollias,
        Eleni~Papadopoulou, Paraskevi~Tzouveli, Jie~Wei,
        and~Xinyu~Yang

\thanks{The work of Guanyu Hu was supported by the Innovation Leadership Talent Development Program of Xi'an Jiaotong University.}
\thanks{Guanyu Hu and Dimitrios~Kollias contributed equally to this work}
\thanks{Guanyu Hu and Xinyu Yang are with the School of Computer Science and Technology, Xi'an Jiaotong University. Eleni Papadopoulou and Paraskevi Tzouveli are with the National Technical University of Athens, Greece. Guanyu Hu and Dimitrios~Kollias are with School of Electronic Engineering and Computer Science, Queen Mary University of London. Jie Wei is with the China Mobile Group Shaanxi Co Ltd.}
\thanks{Corresponding author: Dimitrios~Kollias; email: d.kollias@qmul.ac.uk}
\vspace{-20pt}
}

%
%

\markboth{Journal of \LaTeX\ Class Files,~Vol.~14, No.~8, August~2015}%
{Shell \MakeLowercase{\textit{et al.}}: Bare Demo of IEEEtran.cls for IEEE Journals}
%



\maketitle


\begin{abstract}
Evaluating affect analysis methods presents challenges due to inconsistencies in database partitioning and evaluation protocols, leading to unfair and biased results. Previous studies claim continuous performance improvements, but our findings challenge such assertions. Using these insights, we propose a unified protocol for database partitioning that ensures fairness and comparability. We provide detailed demographic annotations (in terms of race, gender and age), evaluation metrics, and a common framework for expression recognition, action unit detection and valence-arousal estimation. We also rerun the
methods with the new protocol and introduce a new
leaderboards to encourage future research in affect recognition with a fairer comparison.
 Our annotations, code, and pre-trained models are available on \hyperlink{https://github.com/dkollias/Fair-Consistent-Affect-Analysis}{Github}.
\end{abstract}


%
\IEEEpeerreviewmaketitle


\section{Introduction}

%
%
%
%
\IEEEPARstart{I}{n} recent years, the rapid integration of machine/deep learning algorithms into various aspects of our daily lives has heightened awareness of the critical need for fairness and equity in their deployment. As these technologies increasingly influence decision-making processes, it is essential to ensure that their outcomes do not inadvertently reinforce or perpetuate existing societal biases. A key aspect of achieving fairness is examining algorithmic behavior across diverse subpopulation groups, such as age, gender, and race. Addressing potential biases and discriminatory impacts on different demographic cohorts is an ethical imperative, necessitating the development of  models that are not only accurate and efficient but also considerate of the diverse characteristics within our society.

Automatic affect analysis, situated at the intersection of physiology, psychology and AI, has a long history. This field involves automatic: a) recognition of six basic expressions (e.g., anger, happiness), plus neutral state \cite{ekman2002facial}; b) detection of activation of facial action units (AUs), i.e., specific movements of facial muscles \cite{ekman2002facial}; c) estimation of dimensional affect, represented by valence (characterizing an emotional state on a scale from positive to negative) and arousal (characterizing an emotional state on a scale from active to passive) \cite{russell1978evidence}.

Affect analysis methods are typically developed using existing databases with the primary goal of maximizing performance on test sets and surpassing the performance of other methods. Often, this is the sole criterion for evaluation, with little consideration for whether these comparisons are fair or whether the methods are unbiased and perform equally well across subjects of various demographic attributes. This is particularly concerning given that many databases do not have an even distribution of subjects among demographic groups.
Unless explicitly modified, methods are significantly impacted by bias, as they receive more training samples from majority demographic groups in the database. This leads to lower performance for minority groups, i.e., subjects represented by fewer samples, which is undesirable.

Over the past decades, many affective databases have been constructed, evolving from small-scale, controlled environments to large-scale, real-world, unconstrained conditions (termed ``in-the-wild"). Two of the most commonly used in-the-wild databases for Expression Recognition are RAF-DB \cite{li2017reliable} and AffectNet \cite{mollahosseini2017affectnet} (AffectNet is also annotated for valence-arousal). The most widely used databases for AU detection are DISFA \cite{mavadati2013disfa}, GFT \cite{girard2017sayette}, EmotioNet \cite{emotionet2016}, RAF-AU \cite{yan2020raf}. However, these databases present several issues, which we discuss next.

\textit{Some databases, such as DISFA, do not have predefined splits into training, validation, and test sets.} Consequently, many studies perform \textit{k}-fold cross-validation with varying \textit{k} values \cite{mitenkova2019valence,kossaifi2020factorized}, or use the same \textit{k} but different samples in each fold. Additionally, some studies use subject-dependent splits, while others use subject-independent splits (ensuring the same subject appears only in the same split) \cite{parameshwara2023examining,parameshwara2023efficient}. These variations result in non-comparable outcomes, as methods are developed with different training and validation samples, and there are no common test samples for comparison.

\textit{Other databases are split into only two sets.} For example, RAF-DB has training and test sets; AffectNet has training and validation sets (its test set is unpublished); RAF-AU does not provide exact training and test partitions. For RAF-AU, researchers define their own partitions without sharing them, hindering reproducibility \cite{an2024learning}. In some cases, the evaluation set is used for both validation and testing \cite{parameshwara2023efficient,zheng2023poster,li2019self,chaudhari2022vitfer}, or the existing training set is split into new training and validation sets without specifying sample allocations \cite{varsha,handrich2020simultaneous}. This leads to unfair performance comparisons, as methods using more training data outperform those with less training data.

\textit{Some databases consist of both manual and automatic annotations.} For instance, EmotioNet's training set includes only automatic labels, while its validation and test sets have manual labels; AffectNet contains a mix of automatic and manual annotations. Some researchers use the automatic annotations despite their noise, while others split the validation set into new training and validation sets, partitioning differently (e.g., 90-10\% split in \cite{wang2020tal} vs. 80-20\% split in \cite{werner2020facial}) without reporting exact splits, causing inconsistency.

\textit{Databases' test sets are often small, especially compared to the training sets.}  AffectNet's training set has around 290K images, but its test set has only 4K images. Similarly, RAF-DB's training set contains about 12K images, and its test set has only 3K images. Following \cite{shao2024joint}, GFT's training set includes 78 subjects with approximately 110K frames, while the test set has only 18 subjects with about 25K frames.

Evaluating a method's performance on a small test set can lead to inaccurate conclusions. A method that outperforms another on a small test set may not do so on a larger test set. This is proved in our experimental section, where re-partitioning the databases and re-evaluating state-of-the-art methods show differing performance rankings on new partitions.

\textit{Databases have inconsistent annotation forms.} DISFA is annotated with AU intensities (0-5 scale). To convert AU intensities to activation/non-activation, some studies consider intensity \>0 as activation and 0 as non-activation \cite{li2019self,liu2023pose}, while others consider intensity \>1 as activation and 0 or 1 as non-activation \cite{zhang2024multimodal,cui2023biomechanics}. DISFA includes videos from the left and right views of 27 subjects, but many studies only consider one view. Additionally, different combinations of AUs are used in each database and in some databases not all the released AU annotations are used.
DISFA has 12 AU annotations, but studies use only 8; GFT has 32 AU annotations, but studies use only 10. For consistency and cross-database experiments, all AUs should be used.

\textit{Evaluation metrics vary across databases and are not always appropriate.} Studies using RAF-DB primarily report total accuracy and  occasionally average accuracy, while AffectNet studies report total accuracy and $\text{F}_{1}$ score. Total accuracy can be misleading in imbalanced test sets, making $\text{F}_{1}$ score and average accuracy more appropriate. DISFA and GFT studies use $\text{F}_{1}$, RAF-AU studies use AUC-ROC and $\text{F}_{1}$, EmotioNet studies use the mean between average accuracy and $\text{F}_{1}$ score. For VA estimation, AffectNet studies use Concordance Correlation Coefficient (CCC), Pearson Correlation Coefficient (PCC), and Mean Squared Error (MSE).

\textit{Databases often have inconsistent label distributions in each partition, as well as imbalanced and small partitions, leading to poor generalizability.} A critical assumption in AI is that data distributions and labels are consistent across training, validation, and test sets. However, this is often not the case (e.g., AffectNet, EmotioNet), causing label shift and affecting performance, as minimizing the objective on the validation set does not ensure good performance on the test set.

\textit{Databases lack even distribution of demographic attributes or considerations of these in their partitions.}  GFT only includes subjects aged 21-28 but its test set excludes ages 25-27 and has no Asian subjects. AffectNet's test set includes only 27 images of people aged 70 or older. \textit{Most databases do not provide labels for demographic attributes, complicating bias assessment and performance evaluation.}

All things considered, the contributions of this work are:
\begin{itemize}
    \item annotating six affective databases in terms of demographic attributes (age, gender and race);
    
    \item partitioning these affective databases according to a common protocol that we define; in that protocol, we pay particular attention to the demographic attributes and to fairness of evaluations;
    
    \item conducting extensive experimental study of various baseline and state-of-the-art methods in each database using the new protocol; various performance metrics are utilized, including ones measuring fairness and bias.
\end{itemize}

\section{Materials \& Methods}

\noindent In this section, we describe the annotation of the six affective databases with respect to demographic attributes. We also define and describe a common protocol for partitioning these databases, addressing the challenges highlighted in the introduction section and considering the demographic annotations. Finally, we present the new partition of the databases and provide pertinent statistical analyses.

\subsection{Annotation}

We annotated all affective databases (AffectNet, RAF-DB, DISFA, EmotioNet, GFT, RAF-AU) in terms of demographic attributes, including age, gender, and race. It is crucial to distinguish between race and ethnicity, as they are often used interchangeably but represent different categorizations of humans. Race is typically defined by physical traits, while ethnicity is based on cultural similarities \cite{schaefer2008encyclopedia}. For the purposes of our annotations, we focused on race, as it is more readily discernible and annotatable based on facial appearance.
We adopted a widely accepted race classification system from the U.S. Census Bureau, categorizing individuals into the following five race groups: Asian, Black (or African American), Indian (or Alaska Native), Native Hawaiian (or Other Pacific Islander), and White (or Caucasian). During the annotation process, we did not encounter any instances of individuals identified as Native Hawaiian or Other Pacific Islander.


We further annotated all databases in terms of gender, categorizing subjects as Male, Female, or Other/Uncertain. The ``Uncertain" category primarily includes instances where determining gender is challenging, such as infants below the age of 3 or adolescents aged 4-19 with significant obstructions in the central part of the image, like hats or hands.

For age annotation, we divided subjects into nine categories: $\leq$2, 3–9, 10–19, 20–29, 30–39, 40–49, 50–59, 60–69, $\geq$70. This categorization was chosen to: i) align with age ranges in other databases; ii) provide more fine-grained categories rather than broad ones like ‘young,’ ‘middle-aged,’ or ‘old’; iii) facilitate easier and less noisy annotation, as it is more straightforward to determine if a person is between 40 and 49 years old than to pinpoint a specific age within that range.

We developed custom annotation software to independently annotate each image and video in terms of race, gender, and age group. Two annotators independently labelled approximately 20,000 images from each image database and all videos from each video database. When both annotators agreed on their judgments, we adopted these as the ground truth.
The annotation process also included a final step where we trained a model using our ground truth annotations and tested it on the remaining images in each image database so that we create annotations for these images as well.


\subsection{Protocol}
In this subsection, we delineate the protocol for creating new partitions for each of the six affective databases. Each database has been split uniformly, adhering to the same rationale for consistency. The protocol involves: 
i) partitioning into three sets (training, validation, and test); 
ii) ensuring each set contains an adequate amount of data and subjects, to the extent possible given the overall dataset size; 
iii) maintaining similar distributions across affect labels (expressions, AUs, or VA), age groups, race groups, and genders in each set; 
iv) ensuring subject independence across sets; 
v) using only manual annotations for affect labels; 
vi) employing specific performance evaluation metrics; and 
vii) using a consistent annotation format.
To elaborate on these points: 
Point (vi):  $\text{F}_{1}$ score is chosen for evaluating AU Detection and Expression Recognition. For VA estimation, CCC is selected.
Point (vii): In DISFA, all 12 AUs are utilized. AU labels with intensity of $>0$ indicate AU activation, while an intensity of 0 indicates AU non-activation. Both subjects' views are included in the new partition. In GFT, all 14 AUs are used.
For Point (iii), the training, validation, and test sets follow the 55\%-15\%-30\% rule. According to this rule, the training set comprises 55\% of the data (spanning all affect labels, race groups, age groups and genders), the validation set comprises 15\% of the data, and the test set comprises the remaining 30\% of the data.

\begin{figure}[tb]
\centering
  \includegraphics[width=0.4\textwidth]{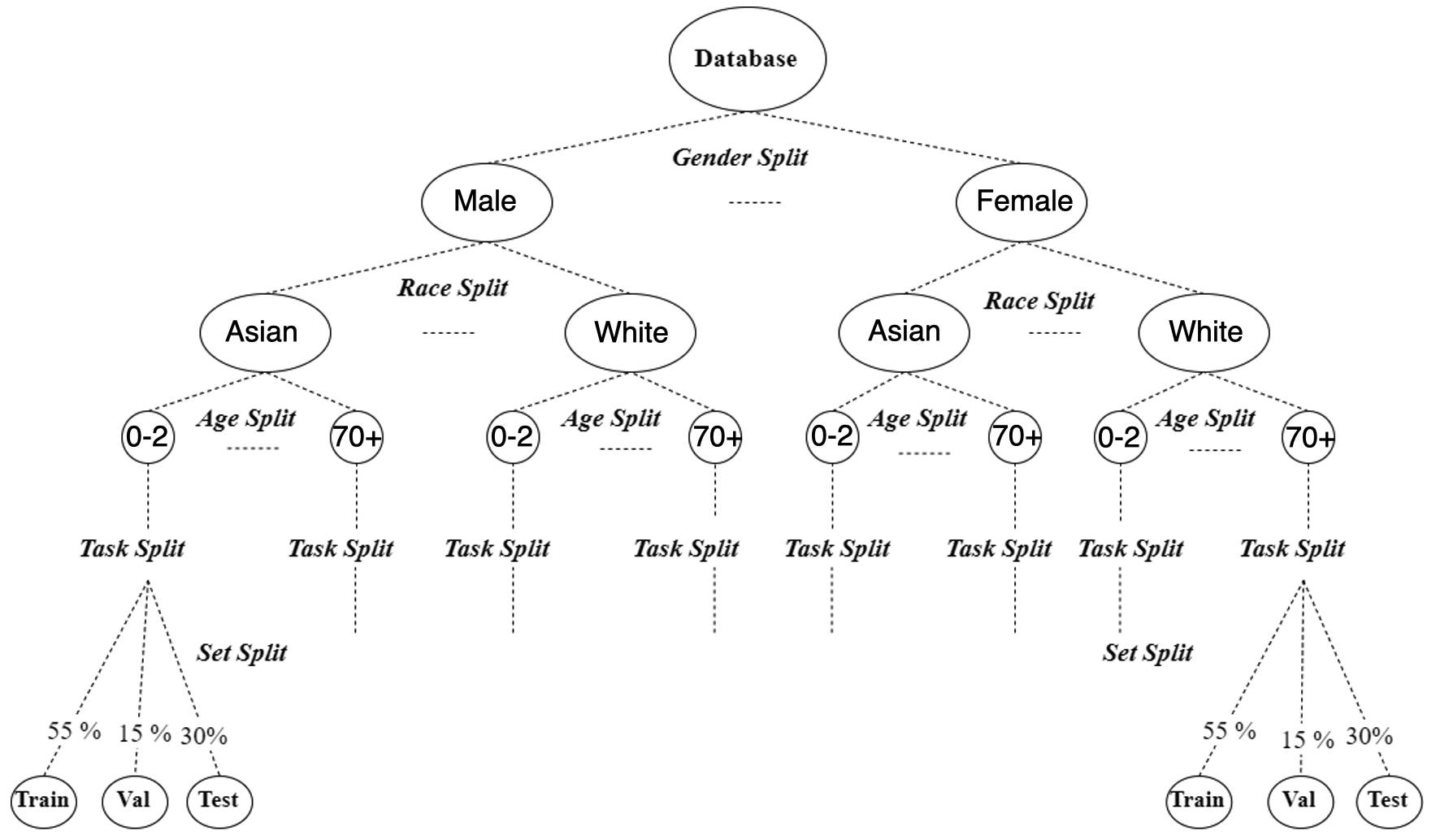}
  \caption{The proposed protocol and subsequent partition of a database}
  \label{protocol1}
\end{figure}

\begin{figure}[h]
\centering
  \includegraphics[width=0.4\textwidth]{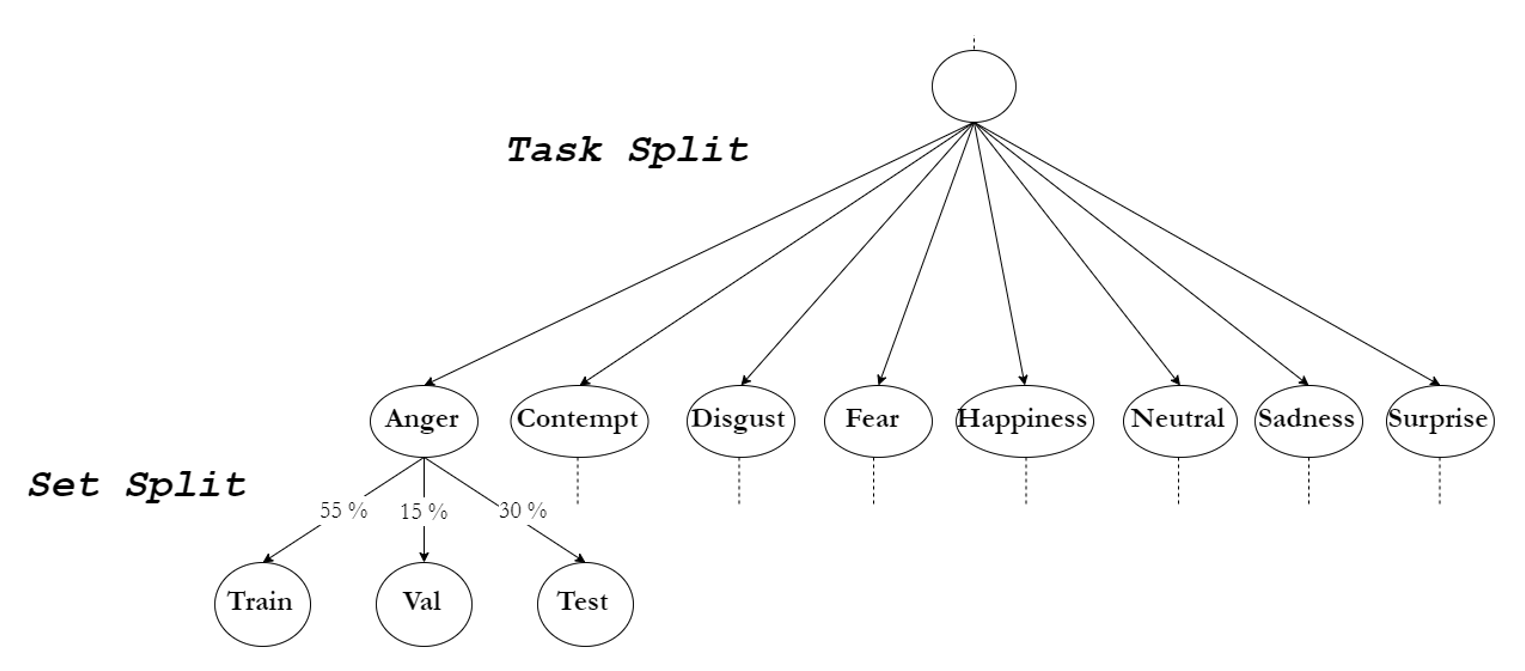}
    \vspace{-10pt}
  
  \caption{`Task Split' part of protocol \& partition in case of basic expressions}
  \label{protocol2}
\end{figure}

\begin{figure}[h]
\centering
  \includegraphics[width=0.4\textwidth]{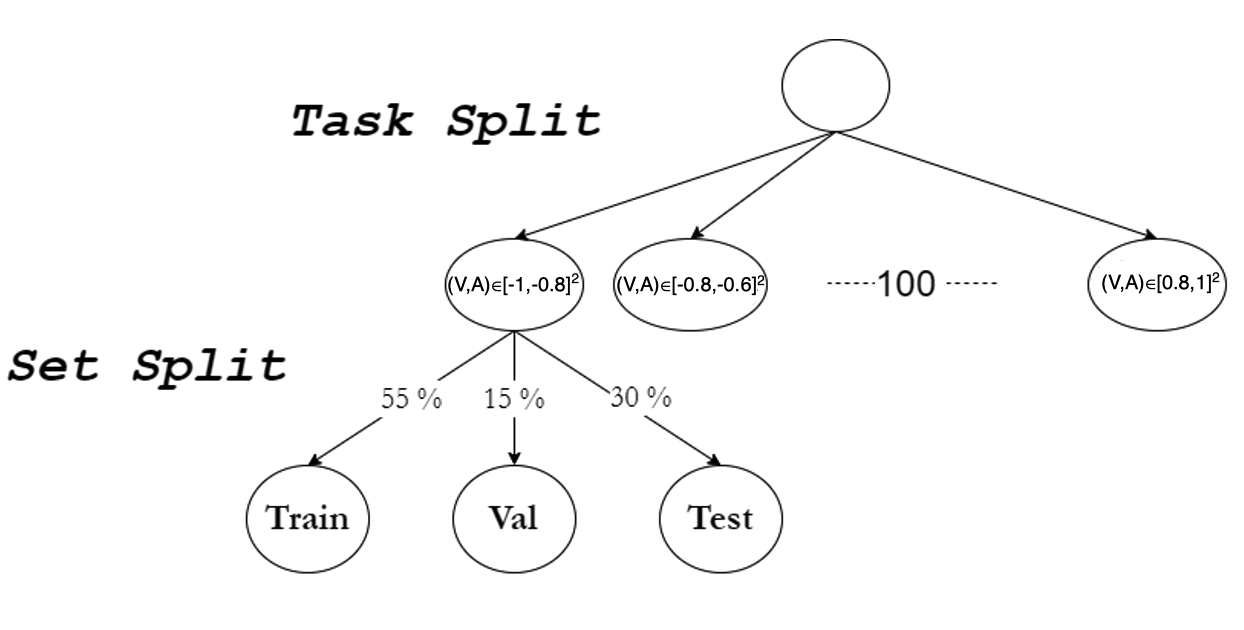}
  \vspace{-10pt}
  \caption{`Task Split' part of proposed protocol and partition in case of VA}
  \label{protocol3}
\end{figure}

Figure \ref{protocol1} illustrates the proposed protocol and database partition. Figure \ref{protocol2} illustrates the `Task Split' part of the proposed protocol and database partition in the case of expressions. In the VA case, at first we convert the continuous 2D space into distinct regions (one such region is when valence takes values in $[-1,-0.8]$ and arousal takes values in $[-0.8,-0.6]$). Figure \ref{protocol3} illustrates the `Task Split' part of the proposed protocol and database partition in the VA case.


\subsection{Databases: Original vs New Partition}

\begin{figure}[h]   
	\centering
	\includegraphics[width=1\linewidth]{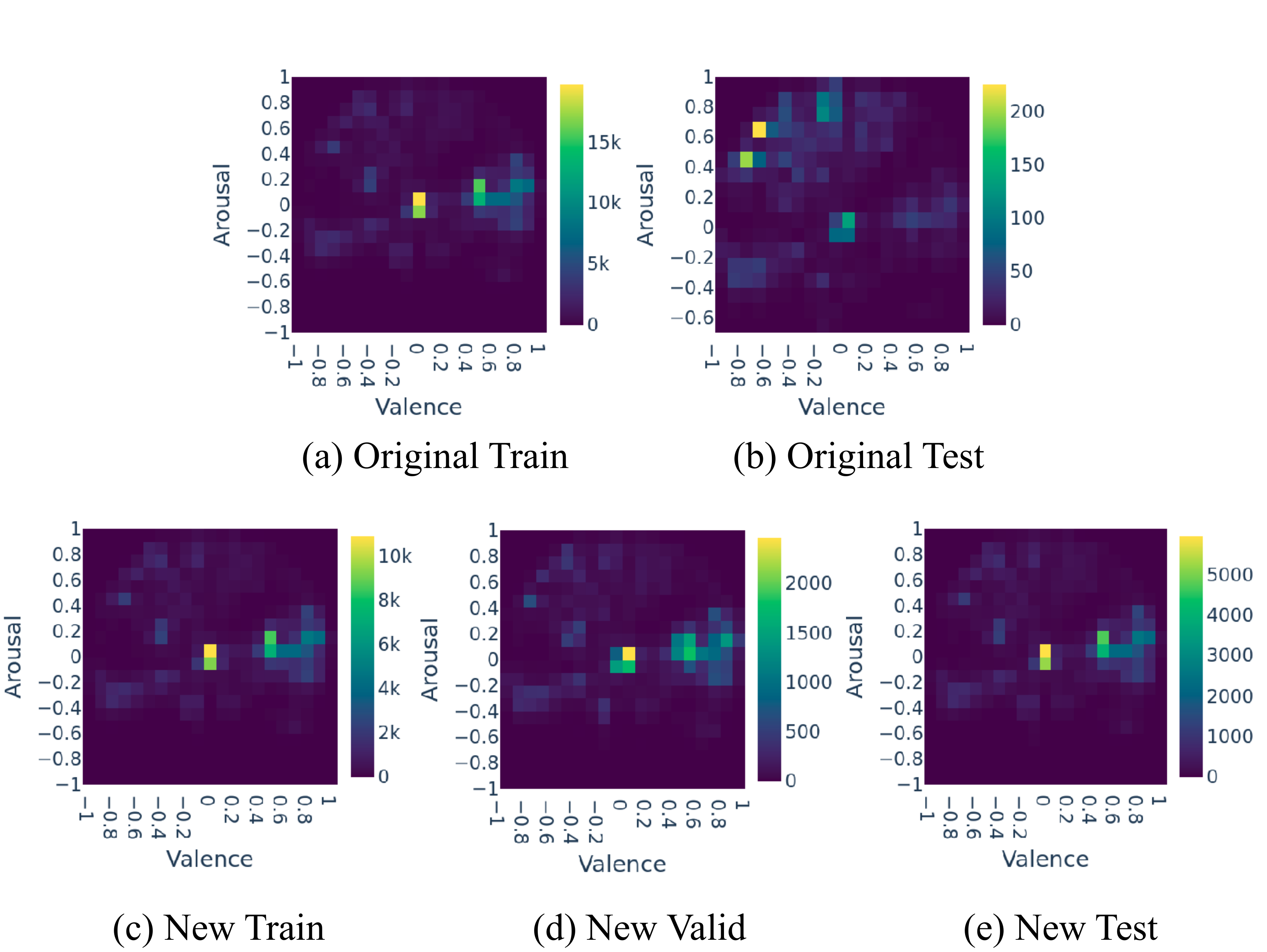}   
	\caption{2D Valence-Arousal Histogram: A Comparison Between the Original and New Partitions of AffectNet}
	\label{fig:fig1} 
\end{figure}

Tables \ref{table1}, \ref{table2} and \ref{table3}, as well as Figure \ref{fig:fig1}  present information (Label, Gender, Race, Age) regarding the original and new partitions (according to the previously mentioned protocol) of all utilized affective databases.

\begin{table}[]
\scriptsize
\caption{Data Statistics for the Original and New Partitions of Databases Annotated in terms of facial expressions }\label{table1}
\setlength{\tabcolsep}{0.7mm}
\renewcommand{\arraystretch}{1.1} 
\scalebox{.92}{
\begin{tabular}{|cc|ccccc|ccccc|}
\hline
\multicolumn{2}{|c|}{\textbf{Database}} &
  \multicolumn{5}{c|}{\textbf{AffectNet}} &
  \multicolumn{5}{c|}{\textbf{RAF-DB}} \\ \hline
\multicolumn{2}{|c|}{\textbf{Partition}} &
  \multicolumn{2}{c|}{\textbf{Original}} &
  \multicolumn{3}{c|}{\textbf{New}} &
  \multicolumn{2}{c|}{\textbf{Original}} &
  \multicolumn{3}{c|}{\textbf{New}} \\ \hline
\multicolumn{2}{|c|}{\textbf{Set}} &
  \multicolumn{1}{c|}{\textbf{Train}} &
  \multicolumn{1}{c|}{\textbf{Test}} &
  \multicolumn{1}{c|}{\textbf{Train}} &
  \multicolumn{1}{c|}{\textbf{Valid}} &
  \textbf{Test} &
  \multicolumn{1}{c|}{\textbf{Train}} &
  \multicolumn{1}{c|}{\textbf{Test}} &
  \multicolumn{1}{c|}{\textbf{Train}} &
  \multicolumn{1}{c|}{\textbf{Valid}} &
  \textbf{Test} \\ \hline
\multicolumn{2}{|c|}{\textbf{Total}} &
  \multicolumn{1}{c|}{287651} &
  \multicolumn{1}{c|}{3999} &
  \multicolumn{1}{c|}{159540} &
  \multicolumn{1}{c|}{43330} &
  87710 &
  \multicolumn{1}{c|}{12271} &
  \multicolumn{1}{c|}{3068} &
  \multicolumn{1}{c|}{8327} &
  \multicolumn{1}{c|}{2206} &
  4806 \\ \hline
\multicolumn{1}{|c|}{\multirow{2}{*}{\textbf{Gen.}}} &
  \textbf{Female} &
  \multicolumn{1}{c|}{144422} &
  \multicolumn{1}{c|}{1794} &
  \multicolumn{1}{c|}{80281} &
  \multicolumn{1}{c|}{21808} &
  44127 &
  \multicolumn{1}{c|}{6562} &
  \multicolumn{1}{c|}{1620} &
  \multicolumn{1}{c|}{4455} &
  \multicolumn{1}{c|}{1189} &
  2538 \\ \cline{2-12} 
\multicolumn{1}{|c|}{} &
  \textbf{Male} &
  \multicolumn{1}{c|}{142172} &
  \multicolumn{1}{c|}{2192} &
  \multicolumn{1}{c|}{79259} &
  \multicolumn{1}{c|}{21522} &
  43583 &
  \multicolumn{1}{c|}{4957} &
  \multicolumn{1}{c|}{1249} &
  \multicolumn{1}{c|}{3366} &
  \multicolumn{1}{c|}{891} &
  1949 \\ \hline
\multicolumn{1}{|c|}{\multirow{4}{*}{\textbf{Race}}} &
  \textbf{Asian} &
  \multicolumn{1}{c|}{25573} &
  \multicolumn{1}{c|}{311} &
  \multicolumn{1}{c|}{14168} &
  \multicolumn{1}{c|}{3823} &
  7893 &
  \multicolumn{1}{c|}{1912} &
  \multicolumn{1}{c|}{483} &
  \multicolumn{1}{c|}{1285} &
  \multicolumn{1}{c|}{329} &
  781 \\ \cline{2-12} 
\multicolumn{1}{|c|}{} &
  \textbf{Black} &
  \multicolumn{1}{c|}{23578} &
  \multicolumn{1}{c|}{321} &
  \multicolumn{1}{c|}{13072} &
  \multicolumn{1}{c|}{3518} &
  7309 &
  \multicolumn{1}{c|}{968} &
  \multicolumn{1}{c|}{234} &
  \multicolumn{1}{c|}{631} &
  \multicolumn{1}{c|}{156} &
  415 \\ \cline{2-12} 
\multicolumn{1}{|c|}{} &
  \textbf{Indian} &
  \multicolumn{1}{c|}{17201} &
  \multicolumn{1}{c|}{234} &
  \multicolumn{1}{c|}{9526} &
  \multicolumn{1}{c|}{2552} &
  5357 &
  \multicolumn{1}{c|}{/} &
  \multicolumn{1}{c|}{/} &
  \multicolumn{1}{c|}{/} &
  \multicolumn{1}{c|}{/} &
  / \\ \cline{2-12} 
\multicolumn{1}{|c|}{} &
  \textbf{White} &
  \multicolumn{1}{c|}{220242} &
  \multicolumn{1}{c|}{3120} &
  \multicolumn{1}{c|}{122774} &
  \multicolumn{1}{c|}{33437} &
  67151 &
  \multicolumn{1}{c|}{9391} &
  \multicolumn{1}{c|}{2351} &
  \multicolumn{1}{c|}{6411} &
  \multicolumn{1}{c|}{1721} &
  3610 \\ \hline
\multicolumn{1}{|c|}{\multirow{9}{*}{\textbf{Age}}} &
  \textbf{$\le$ 2} &
  \multicolumn{1}{c|}{15418} &
  \multicolumn{1}{c|}{288} &
  \multicolumn{1}{c|}{8607} &
  \multicolumn{1}{c|}{2327} &
  4772 &
  \multicolumn{1}{c|}{1283} &
  \multicolumn{1}{c|}{329} &
  \multicolumn{1}{c|}{865} &
  \multicolumn{1}{c|}{219} &
  528 \\ \cline{2-12} 
\multicolumn{1}{|c|}{} &
  \textbf{03-09} &
  \multicolumn{1}{c|}{17794} &
  \multicolumn{1}{c|}{257} &
  \multicolumn{1}{c|}{9895} &
  \multicolumn{1}{c|}{2684} &
  5472 &
  \multicolumn{1}{c|}{\multirow{2}{*}{2171}} &
  \multicolumn{1}{c|}{\multirow{2}{*}{486}} &
  \multicolumn{1}{c|}{\multirow{2}{*}{1436}} &
  \multicolumn{1}{c|}{\multirow{2}{*}{377}} &
  \multirow{2}{*}{844} \\ \cline{2-7}
\multicolumn{1}{|c|}{} &
  \textbf{10-19} &
  \multicolumn{1}{c|}{15042} &
  \multicolumn{1}{c|}{177} &
  \multicolumn{1}{c|}{8337} &
  \multicolumn{1}{c|}{2249} &
  4633 &
  \multicolumn{1}{c|}{} &
  \multicolumn{1}{c|}{} &
  \multicolumn{1}{c|}{} &
  \multicolumn{1}{c|}{} &
   \\ \cline{2-12} 
\multicolumn{1}{|c|}{} &
  \textbf{20-29} &
  \multicolumn{1}{c|}{117914} &
  \multicolumn{1}{c|}{1464} &
  \multicolumn{1}{c|}{65624} &
  \multicolumn{1}{c|}{17881} &
  35873 &
  \multicolumn{1}{c|}{\multirow{2}{*}{6531}} &
  \multicolumn{1}{c|}{\multirow{2}{*}{1662}} &
  \multicolumn{1}{c|}{\multirow{2}{*}{4480}} &
  \multicolumn{1}{c|}{\multirow{2}{*}{1212}} &
  \multirow{2}{*}{2501} \\ \cline{2-7}
\multicolumn{1}{|c|}{} &
  \textbf{30-39} &
  \multicolumn{1}{c|}{56462} &
  \multicolumn{1}{c|}{821} &
  \multicolumn{1}{c|}{31476} &
  \multicolumn{1}{c|}{8564} &
  17243 &
  \multicolumn{1}{c|}{} &
  \multicolumn{1}{c|}{} &
  \multicolumn{1}{c|}{} &
  \multicolumn{1}{c|}{} &
   \\ \cline{2-12} 
\multicolumn{1}{|c|}{} &
  \textbf{40-49} &
  \multicolumn{1}{c|}{28280} &
  \multicolumn{1}{c|}{427} &
  \multicolumn{1}{c|}{15758} &
  \multicolumn{1}{c|}{4272} &
  8677 &
  \multicolumn{1}{c|}{\multirow{3}{*}{1920}} &
  \multicolumn{1}{c|}{\multirow{3}{*}{502}} &
  \multicolumn{1}{c|}{\multirow{3}{*}{1312}} &
  \multicolumn{1}{c|}{\multirow{3}{*}{344}} &
  \multirow{3}{*}{766} \\ \cline{2-7}
\multicolumn{1}{|c|}{} &
  \textbf{50-59} &
  \multicolumn{1}{c|}{22950} &
  \multicolumn{1}{c|}{363} &
  \multicolumn{1}{c|}{12790} &
  \multicolumn{1}{c|}{3467} &
  7056 &
  \multicolumn{1}{c|}{} &
  \multicolumn{1}{c|}{} &
  \multicolumn{1}{c|}{} &
  \multicolumn{1}{c|}{} &
   \\ \cline{2-7}
\multicolumn{1}{|c|}{} &
  \textbf{60-69} &
  \multicolumn{1}{c|}{9395} &
  \multicolumn{1}{c|}{162} &
  \multicolumn{1}{c|}{5227} &
  \multicolumn{1}{c|}{1407} &
  2923 &
  \multicolumn{1}{c|}{} &
  \multicolumn{1}{c|}{} &
  \multicolumn{1}{c|}{} &
  \multicolumn{1}{c|}{} &
   \\ \cline{2-12} 
\multicolumn{1}{|c|}{} &
  \textbf{$\ge$ 70} &
  \multicolumn{1}{c|}{3339} &
  \multicolumn{1}{c|}{27} &
  \multicolumn{1}{c|}{1826} &
  \multicolumn{1}{c|}{479} &
  1061 &
  \multicolumn{1}{c|}{366} &
  \multicolumn{1}{c|}{89} &
  \multicolumn{1}{c|}{234} &
  \multicolumn{1}{c|}{54} &
  167 \\ \hline
\multicolumn{1}{|c|}{\multirow{8}{*}{\textbf{Expr.}}} &
  \textbf{Neutral} &
  \multicolumn{1}{c|}{74874} &
  \multicolumn{1}{c|}{500} &
  \multicolumn{1}{c|}{41252} &
  \multicolumn{1}{c|}{11223} &
  22588 &
  \multicolumn{1}{c|}{2524} &
  \multicolumn{1}{c|}{680} &
  \multicolumn{1}{c|}{1744} &
  \multicolumn{1}{c|}{463} &
  997 \\ \cline{2-12} 
\multicolumn{1}{|c|}{} &
  \textbf{Happiness} &
  \multicolumn{1}{c|}{134415} &
  \multicolumn{1}{c|}{500} &
  \multicolumn{1}{c|}{73958} &
  \multicolumn{1}{c|}{20144} &
  40432 &
  \multicolumn{1}{c|}{4772} &
  \multicolumn{1}{c|}{1185} &
  \multicolumn{1}{c|}{3260} &
  \multicolumn{1}{c|}{877} &
  1820 \\ \cline{2-12} 
\multicolumn{1}{|c|}{} &
  \textbf{Sadness} &
  \multicolumn{1}{c|}{25459} &
  \multicolumn{1}{c|}{500} &
  \multicolumn{1}{c|}{14174} &
  \multicolumn{1}{c|}{3842} &
  7816 &
  \multicolumn{1}{c|}{1982} &
  \multicolumn{1}{c|}{478} &
  \multicolumn{1}{c|}{1332} &
  \multicolumn{1}{c|}{353} &
  775 \\ \cline{2-12} 
\multicolumn{1}{|c|}{} &
  \textbf{Surprise} &
  \multicolumn{1}{c|}{14090} &
  \multicolumn{1}{c|}{500} &
  \multicolumn{1}{c|}{7951} &
  \multicolumn{1}{c|}{2148} &
  4425 &
  \multicolumn{1}{c|}{1290} &
  \multicolumn{1}{c|}{329} &
  \multicolumn{1}{c|}{872} &
  \multicolumn{1}{c|}{229} &
  518 \\ \cline{2-12} 
\multicolumn{1}{|c|}{} &
  \textbf{Fear} &
  \multicolumn{1}{c|}{6378} &
  \multicolumn{1}{c|}{500} &
  \multicolumn{1}{c|}{3729} &
  \multicolumn{1}{c|}{997} &
  2113 &
  \multicolumn{1}{c|}{281} &
  \multicolumn{1}{c|}{74} &
  \multicolumn{1}{c|}{184} &
  \multicolumn{1}{c|}{44} &
  127 \\ \cline{2-12} 
\multicolumn{1}{|c|}{} &
  \textbf{Disgust} &
  \multicolumn{1}{c|}{3803} &
  \multicolumn{1}{c|}{500} &
  \multicolumn{1}{c|}{2318} &
  \multicolumn{1}{c|}{611} &
  1355 &
  \multicolumn{1}{c|}{717} &
  \multicolumn{1}{c|}{160} &
  \multicolumn{1}{c|}{471} &
  \multicolumn{1}{c|}{119} &
  287 \\ \cline{2-12} 
\multicolumn{1}{|c|}{} &
  \textbf{Anger} &
  \multicolumn{1}{c|}{24882} &
  \multicolumn{1}{c|}{500} &
  \multicolumn{1}{c|}{13859} &
  \multicolumn{1}{c|}{3755} &
  7649 &
  \multicolumn{1}{c|}{705} &
  \multicolumn{1}{c|}{162} &
  \multicolumn{1}{c|}{464} &
  \multicolumn{1}{c|}{121} &
  282 \\ \cline{2-12} 
\multicolumn{1}{|c|}{} &
  \textbf{Contempt} &
  \multicolumn{1}{c|}{3750} &
  \multicolumn{1}{c|}{499} &
  \multicolumn{1}{c|}{2299} &
  \multicolumn{1}{c|}{610} &
  1332 &
  \multicolumn{1}{c|}{/} &
  \multicolumn{1}{c|}{/} &
  \multicolumn{1}{c|}{/} &
  \multicolumn{1}{c|}{/} &
  / \\ \hline
\end{tabular}}
\end{table}

\begin{table*}[]
\centering\scriptsize
\caption{Data Statistics for the Original  and New Partitions of Databases Annotated for AUs (`Org.' denotes Original partition)}\label{table2}
\setlength{\tabcolsep}{1mm}
\renewcommand{\arraystretch}{1} 
\scalebox{0.95}{
\begin{tabular}{|cc|cccc|ccccc|ccccc|cccc|}
\hline
\multicolumn{2}{|c|}{\textbf{Database}} &
  \multicolumn{4}{c|}{\textbf{DISFA}} &
  \multicolumn{5}{c|}{\textbf{EmotioNet}} &
  \multicolumn{5}{c|}{\textbf{GFT}} &
  \multicolumn{4}{c|}{\textbf{RAF-AU}} \\ \hline
\multicolumn{2}{|c|}{\textbf{Partition}} &
  \multicolumn{1}{c|}{\multirow{2}{*}{\textbf{Org.}}} &
  \multicolumn{3}{c|}{\textbf{New}} &
  \multicolumn{2}{c|}{\textbf{Org.}} &
  \multicolumn{3}{c|}{\textbf{New}} &
  \multicolumn{2}{c|}{\textbf{Org.}} &
  \multicolumn{3}{c|}{\textbf{New}} &
  \multicolumn{1}{c|}{\multirow{2}{*}{\textbf{Org.}}} &
  \multicolumn{3}{c|}{\textbf{New}} \\ \cline{1-2} \cline{4-16} \cline{18-20} 
\multicolumn{2}{|c|}{\textbf{Set}} &
  \multicolumn{1}{c|}{} &
  \multicolumn{1}{c|}{\textbf{Train}} &
  \multicolumn{1}{c|}{\textbf{Valid}} &
  \textbf{Test} &
  \multicolumn{1}{c|}{\textbf{Train}} &
  \multicolumn{1}{c|}{\textbf{Test}} &
  \multicolumn{1}{c|}{\textbf{Train}} &
  \multicolumn{1}{c|}{\textbf{Valid}} &
  \textbf{Test} &
  \multicolumn{1}{c|}{\textbf{Train}} &
  \multicolumn{1}{c|}{\textbf{Test}} &
  \multicolumn{1}{c|}{\textbf{Train}} &
  \multicolumn{1}{c|}{\textbf{Valid}} &
  \textbf{Test} &
  \multicolumn{1}{c|}{} &
  \multicolumn{1}{c|}{\textbf{Train}} &
  \multicolumn{1}{c|}{\textbf{Valid}} &
  \textbf{Test} \\ \hline
\multicolumn{2}{|c|}{\textbf{Total Amount}} &
  \multicolumn{1}{c|}{261628} &
  \multicolumn{1}{c|}{141041} &
  \multicolumn{1}{c|}{29070} &
  84442 &
  \multicolumn{1}{c|}{24644} &
  \multicolumn{1}{c|}{20361} &
  \multicolumn{1}{c|}{24674} &
  \multicolumn{1}{c|}{6705} &
  13549 &
  \multicolumn{1}{c|}{108549} &
  \multicolumn{1}{c|}{24645} &
  \multicolumn{1}{c|}{69618} &
  \multicolumn{1}{c|}{23685} &
  39891 &
  \multicolumn{1}{c|}{4601} &
  \multicolumn{1}{c|}{2469} &
  \multicolumn{1}{c|}{645} &
  1426 \\ \hline
\multicolumn{1}{|c|}{\multirow{2}{*}{\textbf{Gender}}} &
  \textbf{Female} &
  \multicolumn{1}{c|}{116278} &
  \multicolumn{1}{c|}{63521} &
  \multicolumn{1}{c|}{9690} &
  38723 &
  \multicolumn{1}{c|}{11716} &
  \multicolumn{1}{c|}{10132} &
  \multicolumn{1}{c|}{11996} &
  \multicolumn{1}{c|}{3261} &
  6591 &
  \multicolumn{1}{c|}{45736} &
  \multicolumn{1}{c|}{11120} &
  \multicolumn{1}{c|}{30452} &
  \multicolumn{1}{c|}{9778} &
  16626 &
  \multicolumn{1}{c|}{2301} &
  \multicolumn{1}{c|}{1250} &
  \multicolumn{1}{c|}{328} &
  723 \\ \cline{2-20} 
\multicolumn{1}{|c|}{} &
  \textbf{Male} &
  \multicolumn{1}{c|}{145350} &
  \multicolumn{1}{c|}{77520} &
  \multicolumn{1}{c|}{19380} &
  45719 &
  \multicolumn{1}{c|}{12892} &
  \multicolumn{1}{c|}{10188} &
  \multicolumn{1}{c|}{12678} &
  \multicolumn{1}{c|}{3444} &
  6958 &
  \multicolumn{1}{c|}{62813} &
  \multicolumn{1}{c|}{13525} &
  \multicolumn{1}{c|}{39166} &
  \multicolumn{1}{c|}{13907} &
  23265 &
  \multicolumn{1}{c|}{2239} &
  \multicolumn{1}{c|}{1219} &
  \multicolumn{1}{c|}{317} &
  703 \\ \hline
\multicolumn{1}{|c|}{\multirow{4}{*}{\textbf{Race}}} &
  \textbf{Asian} &
  \multicolumn{1}{c|}{29070} &
  \multicolumn{1}{c|}{9690} &
  \multicolumn{1}{c|}{/} &
  19296 &
  \multicolumn{1}{c|}{1410} &
  \multicolumn{1}{c|}{1043} &
  \multicolumn{1}{c|}{1340} &
  \multicolumn{1}{c|}{359} &
  754 &
  \multicolumn{1}{c|}{2641} &
  \multicolumn{1}{c|}{/} &
  \multicolumn{1}{c|}{/} &
  \multicolumn{1}{c|}{/} &
  2641 &
  \multicolumn{1}{c|}{453} &
  \multicolumn{1}{c|}{243} &
  \multicolumn{1}{c|}{58} &
  152 \\ \cline{2-20} 
\multicolumn{1}{|c|}{} &
  \textbf{Black} &
  \multicolumn{1}{c|}{9690} &
  \multicolumn{1}{c|}{/} &
  \multicolumn{1}{c|}{/} &
  9690 &
  \multicolumn{1}{c|}{1979} &
  \multicolumn{1}{c|}{1935} &
  \multicolumn{1}{c|}{2144} &
  \multicolumn{1}{c|}{579} &
  1191 &
  \multicolumn{1}{c|}{7989} &
  \multicolumn{1}{c|}{4891} &
  \multicolumn{1}{c|}{9050} &
  \multicolumn{1}{c|}{/} &
  3830 &
  \multicolumn{1}{c|}{274} &
  \multicolumn{1}{c|}{144} &
  \multicolumn{1}{c|}{34} &
  96 \\ \cline{2-20} 
\multicolumn{1}{|c|}{} &
  \textbf{Indian} &
  \multicolumn{1}{c|}{9690} &
  \multicolumn{1}{c|}{/} &
  \multicolumn{1}{c|}{/} &
  9690 &
  \multicolumn{1}{c|}{1442} &
  \multicolumn{1}{c|}{1210} &
  \multicolumn{1}{c|}{1450} &
  \multicolumn{1}{c|}{388} &
  814 &
  \multicolumn{1}{c|}{/} &
  \multicolumn{1}{c|}{/} &
  \multicolumn{1}{c|}{/} &
  \multicolumn{1}{c|}{/} &
  / &
  \multicolumn{1}{c|}{252} &
  \multicolumn{1}{c|}{131} &
  \multicolumn{1}{c|}{29} &
  92 \\ \cline{2-20} 
\multicolumn{1}{|c|}{} &
  \textbf{White} &
  \multicolumn{1}{c|}{213178} &
  \multicolumn{1}{c|}{131351} &
  \multicolumn{1}{c|}{29070} &
  45766 &
  \multicolumn{1}{c|}{19777} &
  \multicolumn{1}{c|}{16132} &
  \multicolumn{1}{c|}{19740} &
  \multicolumn{1}{c|}{5379} &
  10790 &
  \multicolumn{1}{c|}{97919} &
  \multicolumn{1}{c|}{18505} &
  \multicolumn{1}{c|}{60568} &
  \multicolumn{1}{c|}{23685} &
  32171 &
  \multicolumn{1}{c|}{3561} &
  \multicolumn{1}{c|}{1951} &
  \multicolumn{1}{c|}{524} &
  1086 \\ \hline
\multicolumn{1}{|c|}{\multirow{9}{*}{\textbf{Age}}} &
  \textbf{$ \leq $ 2} &
  \multicolumn{1}{c|}{/} &
  \multicolumn{1}{c|}{/} &
  \multicolumn{1}{c|}{/} &
  / &
  \multicolumn{1}{c|}{583} &
  \multicolumn{1}{c|}{262} &
  \multicolumn{1}{c|}{460} &
  \multicolumn{1}{c|}{123} &
  262 &
  \multicolumn{1}{c|}{/} &
  \multicolumn{1}{c|}{/} &
  \multicolumn{1}{c|}{/} &
  \multicolumn{1}{c|}{/} &
  / &
  \multicolumn{1}{c|}{352} &
  \multicolumn{1}{c|}{191} &
  \multicolumn{1}{c|}{48} &
  113 \\ \cline{2-20} 
\multicolumn{1}{|c|}{} &
  \textbf{03-09} &
  \multicolumn{1}{c|}{/} &
  \multicolumn{1}{c|}{/} &
  \multicolumn{1}{c|}{/} &
  / &
  \multicolumn{1}{c|}{939} &
  \multicolumn{1}{c|}{488} &
  \multicolumn{1}{c|}{782} &
  \multicolumn{1}{c|}{210} &
  435 &
  \multicolumn{1}{c|}{/} &
  \multicolumn{1}{c|}{/} &
  \multicolumn{1}{c|}{/} &
  \multicolumn{1}{c|}{/} &
  / &
  \multicolumn{1}{c|}{585} &
  \multicolumn{1}{c|}{318} &
  \multicolumn{1}{c|}{83} &
  184 \\ \cline{2-20} 
\multicolumn{1}{|c|}{} &
  \textbf{10-19} &
  \multicolumn{1}{c|}{9690} &
  \multicolumn{1}{c|}{/} &
  \multicolumn{1}{c|}{/} &
  9690 &
  \multicolumn{1}{c|}{1047} &
  \multicolumn{1}{c|}{754} &
  \multicolumn{1}{c|}{986} &
  \multicolumn{1}{c|}{266} &
  549 &
  \multicolumn{1}{c|}{/} &
  \multicolumn{1}{c|}{/} &
  \multicolumn{1}{c|}{/} &
  \multicolumn{1}{c|}{/} &
  / &
  \multicolumn{1}{c|}{222} &
  \multicolumn{1}{c|}{117} &
  \multicolumn{1}{c|}{30} &
  75 \\ \cline{2-20} 
\multicolumn{1}{|c|}{} &
  \textbf{20-29} &
  \multicolumn{1}{c|}{213178} &
  \multicolumn{1}{c|}{131351} &
  \multicolumn{1}{c|}{29070} &
  48413 &
  \multicolumn{1}{c|}{10101} &
  \multicolumn{1}{c|}{8912} &
  \multicolumn{1}{c|}{10454} &
  \multicolumn{1}{c|}{2849} &
  5710 &
  \multicolumn{1}{c|}{108549} &
  \multicolumn{1}{c|}{24645} &
  \multicolumn{1}{c|}{69618} &
  \multicolumn{1}{c|}{23685} &
  39891 &
  \multicolumn{1}{c|}{1662} &
  \multicolumn{1}{c|}{912} &
  \multicolumn{1}{c|}{246} &
  504 \\ \cline{2-20} 
\multicolumn{1}{|c|}{} &
  \textbf{30-39} &
  \multicolumn{1}{c|}{19380} &
  \multicolumn{1}{c|}{/} &
  \multicolumn{1}{c|}{/} &
  19333 &
  \multicolumn{1}{c|}{5026} &
  \multicolumn{1}{c|}{4206} &
  \multicolumn{1}{c|}{5075} &
  \multicolumn{1}{c|}{1380} &
  2777 &
  \multicolumn{1}{c|}{/} &
  \multicolumn{1}{c|}{/} &
  \multicolumn{1}{c|}{/} &
  \multicolumn{1}{c|}{/} &
  / &
  \multicolumn{1}{c|}{1068} &
  \multicolumn{1}{c|}{584} &
  \multicolumn{1}{c|}{156} &
  328 \\ \cline{2-20} 
\multicolumn{1}{|c|}{} &
  \textbf{40-49} &
  \multicolumn{1}{c|}{19380} &
  \multicolumn{1}{c|}{9690} &
  \multicolumn{1}{c|}{/} &
  7006 &
  \multicolumn{1}{c|}{3128} &
  \multicolumn{1}{c|}{2581} &
  \multicolumn{1}{c|}{3136} &
  \multicolumn{1}{c|}{852} &
  1721 &
  \multicolumn{1}{c|}{/} &
  \multicolumn{1}{c|}{/} &
  \multicolumn{1}{c|}{/} &
  \multicolumn{1}{c|}{/} &
  / &
  \multicolumn{1}{c|}{385} &
  \multicolumn{1}{c|}{208} &
  \multicolumn{1}{c|}{53} &
  124 \\ \cline{2-20} 
\multicolumn{1}{|c|}{} &
  \textbf{50-59} &
  \multicolumn{1}{c|}{/} &
  \multicolumn{1}{c|}{/} &
  \multicolumn{1}{c|}{/} &
  / &
  \multicolumn{1}{c|}{2484} &
  \multicolumn{1}{c|}{2093} &
  \multicolumn{1}{c|}{2513} &
  \multicolumn{1}{c|}{683} &
  1381 &
  \multicolumn{1}{c|}{/} &
  \multicolumn{1}{c|}{/} &
  \multicolumn{1}{c|}{/} &
  \multicolumn{1}{c|}{/} &
  / &
  \multicolumn{1}{c|}{161} &
  \multicolumn{1}{c|}{86} &
  \multicolumn{1}{c|}{20} &
  55 \\ \cline{2-20} 
\multicolumn{1}{|c|}{} &
  \textbf{60-69} &
  \multicolumn{1}{c|}{/} &
  \multicolumn{1}{c|}{/} &
  \multicolumn{1}{c|}{/} &
  / &
  \multicolumn{1}{c|}{1041} &
  \multicolumn{1}{c|}{834} &
  \multicolumn{1}{c|}{1026} &
  \multicolumn{1}{c|}{278} &
  571 &
  \multicolumn{1}{c|}{/} &
  \multicolumn{1}{c|}{/} &
  \multicolumn{1}{c|}{/} &
  \multicolumn{1}{c|}{/} &
  / &
  \multicolumn{1}{c|}{55} &
  \multicolumn{1}{c|}{27} &
  \multicolumn{1}{c|}{5} &
  23 \\ \cline{2-20} 
\multicolumn{1}{|c|}{} &
  \textbf{$\geq $ 70} &
  \multicolumn{1}{c|}{/} &
  \multicolumn{1}{c|}{/} &
  \multicolumn{1}{c|}{/} &
  / &
  \multicolumn{1}{c|}{259} &
  \multicolumn{1}{c|}{190} &
  \multicolumn{1}{c|}{242} &
  \multicolumn{1}{c|}{64} &
  143 &
  \multicolumn{1}{c|}{/} &
  \multicolumn{1}{c|}{/} &
  \multicolumn{1}{c|}{/} &
  \multicolumn{1}{c|}{/} &
  / &
  \multicolumn{1}{c|}{50} &
  \multicolumn{1}{c|}{26} &
  \multicolumn{1}{c|}{4} &
  20 \\ \hline
\multicolumn{2}{|c|}{\textbf{AU1}} &
  \multicolumn{1}{c|}{17556} &
  \multicolumn{1}{c|}{9425} &
  \multicolumn{1}{c|}{2082} &
  5868 &
  \multicolumn{1}{c|}{1452} &
  \multicolumn{1}{c|}{1170} &
  \multicolumn{1}{c|}{1973} &
  \multicolumn{1}{c|}{371} &
  587 &
  \multicolumn{1}{c|}{4011} &
  \multicolumn{1}{c|}{23540} &
  \multicolumn{1}{c|}{2999} &
  \multicolumn{1}{c|}{431} &
  1686 &
  \multicolumn{1}{c|}{1076} &
  \multicolumn{1}{c|}{595} &
  \multicolumn{1}{c|}{164} &
  307 \\ \hline
\multicolumn{2}{|c|}{\textbf{AU2}} &
  \multicolumn{1}{c|}{14728} &
  \multicolumn{1}{c|}{7482} &
  \multicolumn{1}{c|}{512} &
  6480 &
  \multicolumn{1}{c|}{689} &
  \multicolumn{1}{c|}{713} &
  \multicolumn{1}{c|}{1051} &
  \multicolumn{1}{c|}{247} &
  383 &
  \multicolumn{1}{c|}{14527} &
  \multicolumn{1}{c|}{3009} &
  \multicolumn{1}{c|}{8817} &
  \multicolumn{1}{c|}{2920} &
  5799 &
  \multicolumn{1}{c|}{795} &
  \multicolumn{1}{c|}{412} &
  \multicolumn{1}{c|}{121} &
  254 \\ \hline
\multicolumn{2}{|c|}{\textbf{AU4}} &
  \multicolumn{1}{c|}{49188} &
  \multicolumn{1}{c|}{25692} &
  \multicolumn{1}{c|}{3118} &
  19623 &
  \multicolumn{1}{c|}{2857} &
  \multicolumn{1}{c|}{610} &
  \multicolumn{1}{c|}{3941} &
  \multicolumn{1}{c|}{675} &
  1207 &
  \multicolumn{1}{c|}{3989} &
  \multicolumn{1}{c|}{836} &
  \multicolumn{1}{c|}{3671} &
  \multicolumn{1}{c|}{466} &
  688 &
  \multicolumn{1}{c|}{1817} &
  \multicolumn{1}{c|}{994} &
  \multicolumn{1}{c|}{235} &
  576 \\ \hline
\multicolumn{2}{|c|}{\textbf{AU5}} &
  \multicolumn{1}{c|}{5458} &
  \multicolumn{1}{c|}{4081} &
  \multicolumn{1}{c|}{302} &
  954 &
  \multicolumn{1}{c|}{875} &
  \multicolumn{1}{c|}{1287} &
  \multicolumn{1}{c|}{1252} &
  \multicolumn{1}{c|}{243} &
  424 &
  \multicolumn{1}{c|}{2600} &
  \multicolumn{1}{c|}{397} &
  \multicolumn{1}{c|}{1214} &
  \multicolumn{1}{c|}{166} &
  1617 &
  \multicolumn{1}{c|}{985} &
  \multicolumn{1}{c|}{599} &
  \multicolumn{1}{c|}{140} &
  225 \\ \hline
\multicolumn{2}{|c|}{\textbf{AU6}} &
  \multicolumn{1}{c|}{38968} &
  \multicolumn{1}{c|}{24160} &
  \multicolumn{1}{c|}{2866} &
  11067 &
  \multicolumn{1}{c|}{4572} &
  \multicolumn{1}{c|}{4777} &
  \multicolumn{1}{c|}{7289} &
  \multicolumn{1}{c|}{1395} &
  2809 &
  \multicolumn{1}{c|}{30787} &
  \multicolumn{1}{c|}{6826} &
  \multicolumn{1}{c|}{20954} &
  \multicolumn{1}{c|}{7037} &
  9622 &
  \multicolumn{1}{c|}{450} &
  \multicolumn{1}{c|}{218} &
  \multicolumn{1}{c|}{60} &
  166 \\ \hline
\multicolumn{2}{|c|}{\textbf{AU7}} &
  \multicolumn{1}{c|}{/} &
  \multicolumn{1}{c|}{/} &
  \multicolumn{1}{c|}{/} &
  / &
  \multicolumn{1}{c|}{/} &
  \multicolumn{1}{c|}{/} &
  \multicolumn{1}{c|}{/} &
  \multicolumn{1}{c|}{/} &
  / &
  \multicolumn{1}{c|}{49403} &
  \multicolumn{1}{c|}{11108} &
  \multicolumn{1}{c|}{34042} &
  \multicolumn{1}{c|}{10580} &
  15889 &
  \multicolumn{1}{c|}{/} &
  \multicolumn{1}{c|}{/} &
  \multicolumn{1}{c|}{/} &
  / \\ \hline
\multicolumn{2}{|c|}{\textbf{AU9}} &
  \multicolumn{1}{c|}{14264} &
  \multicolumn{1}{c|}{7510} &
  \multicolumn{1}{c|}{1668} &
  4606 &
  \multicolumn{1}{c|}{505} &
  \multicolumn{1}{c|}{143} &
  \multicolumn{1}{c|}{494} &
  \multicolumn{1}{c|}{62} &
  34 &
  \multicolumn{1}{c|}{1519} &
  \multicolumn{1}{c|}{24267} &
  \multicolumn{1}{c|}{1289} &
  \multicolumn{1}{c|}{95} &
  513 &
  \multicolumn{1}{c|}{774} &
  \multicolumn{1}{c|}{385} &
  \multicolumn{1}{c|}{88} &
  294 \\ \hline
\multicolumn{2}{|c|}{\textbf{AU10}} &
  \multicolumn{1}{c|}{/} &
  \multicolumn{1}{c|}{/} &
  \multicolumn{1}{c|}{/} &
  / &
  \multicolumn{1}{c|}{/} &
  \multicolumn{1}{c|}{/} &
  \multicolumn{1}{c|}{/} &
  \multicolumn{1}{c|}{/} &
  / &
  \multicolumn{1}{c|}{26740} &
  \multicolumn{1}{c|}{6079} &
  \multicolumn{1}{c|}{19032} &
  \multicolumn{1}{c|}{5786} &
  8001 &
  \multicolumn{1}{c|}{1390} &
  \multicolumn{1}{c|}{661} &
  \multicolumn{1}{c|}{205} &
  509 \\ \hline
\multicolumn{2}{|c|}{\textbf{AU11}} &
  \multicolumn{1}{c|}{/} &
  \multicolumn{1}{c|}{/} &
  \multicolumn{1}{c|}{/} &
  / &
  \multicolumn{1}{c|}{/} &
  \multicolumn{1}{c|}{/} &
  \multicolumn{1}{c|}{/} &
  \multicolumn{1}{c|}{/} &
  / &
  \multicolumn{1}{c|}{14926} &
  \multicolumn{1}{c|}{3710} &
  \multicolumn{1}{c|}{11117} &
  \multicolumn{1}{c|}{3022} &
  4497 &
  \multicolumn{1}{c|}{/} &
  \multicolumn{1}{c|}{/} &
  \multicolumn{1}{c|}{/} &
  / \\ \hline
\multicolumn{2}{|c|}{\textbf{AU12}} &
  \multicolumn{1}{c|}{61588} &
  \multicolumn{1}{c|}{33754} &
  \multicolumn{1}{c|}{8828} &
  18233 &
  \multicolumn{1}{c|}{7546} &
  \multicolumn{1}{c|}{6641} &
  \multicolumn{1}{c|}{12430} &
  \multicolumn{1}{c|}{2397} &
  5009 &
  \multicolumn{1}{c|}{32021} &
  \multicolumn{1}{c|}{7186} &
  \multicolumn{1}{c|}{21578} &
  \multicolumn{1}{c|}{7426} &
  10203 &
  \multicolumn{1}{c|}{1268} &
  \multicolumn{1}{c|}{584} &
  \multicolumn{1}{c|}{178} &
  488 \\ \hline
\multicolumn{2}{|c|}{\textbf{AU15}} &
  \multicolumn{1}{c|}{15724} &
  \multicolumn{1}{c|}{8631} &
  \multicolumn{1}{c|}{1878} &
  5206 &
  \multicolumn{1}{c|}{/} &
  \multicolumn{1}{c|}{/} &
  \multicolumn{1}{c|}{/} &
  \multicolumn{1}{c|}{/} &
  / &
  \multicolumn{1}{c|}{11536} &
  \multicolumn{1}{c|}{2350} &
  \multicolumn{1}{c|}{8617} &
  \multicolumn{1}{c|}{2231} &
  3038 &
  \multicolumn{1}{c|}{/} &
  \multicolumn{1}{c|}{/} &
  \multicolumn{1}{c|}{/} &
  / \\ \hline
\multicolumn{2}{|c|}{\textbf{AU16}} &
  \multicolumn{1}{c|}{/} &
  \multicolumn{1}{c|}{/} &
  \multicolumn{1}{c|}{/} &
  / &
  \multicolumn{1}{c|}{/} &
  \multicolumn{1}{c|}{/} &
  \multicolumn{1}{c|}{/} &
  \multicolumn{1}{c|}{/} &
  / &
  \multicolumn{1}{c|}{/} &
  \multicolumn{1}{c|}{/} &
  \multicolumn{1}{c|}{/} &
  \multicolumn{1}{c|}{/} &
  / &
  \multicolumn{1}{c|}{720} &
  \multicolumn{1}{c|}{302} &
  \multicolumn{1}{c|}{107} &
  299 \\ \hline
\multicolumn{2}{|c|}{\textbf{AU17}} &
  \multicolumn{1}{c|}{25860} &
  \multicolumn{1}{c|}{12127} &
  \multicolumn{1}{c|}{2290} &
  10931 &
  \multicolumn{1}{c|}{492} &
  \multicolumn{1}{c|}{184} &
  \multicolumn{1}{c|}{515} &
  \multicolumn{1}{c|}{57} &
  58 &
  \multicolumn{1}{c|}{32723} &
  \multicolumn{1}{c|}{7984} &
  \multicolumn{1}{c|}{20395} &
  \multicolumn{1}{c|}{6502} &
  13810 &
  \multicolumn{1}{c|}{541} &
  \multicolumn{1}{c|}{321} &
  \multicolumn{1}{c|}{85} &
  135 \\ \hline
\multicolumn{2}{|c|}{\textbf{AU20}} &
  \multicolumn{1}{c|}{9064} &
  \multicolumn{1}{c|}{5216} &
  \multicolumn{1}{c|}{484} &
  2909 &
  \multicolumn{1}{c|}{134} &
  \multicolumn{1}{c|}{146} &
  \multicolumn{1}{c|}{132} &
  \multicolumn{1}{c|}{17} &
  7 &
  \multicolumn{1}{c|}{/} &
  \multicolumn{1}{c|}{/} &
  \multicolumn{1}{c|}{/} &
  \multicolumn{1}{c|}{/} &
  / &
  \multicolumn{1}{c|}{/} &
  \multicolumn{1}{c|}{/} &
  \multicolumn{1}{c|}{/} &
  / \\ \hline
\multicolumn{2}{|c|}{\textbf{AU23}} &
  \multicolumn{1}{c|}{/} &
  \multicolumn{1}{c|}{/} &
  \multicolumn{1}{c|}{/} &
  / &
  \multicolumn{1}{c|}{/} &
  \multicolumn{1}{c|}{/} &
  \multicolumn{1}{c|}{/} &
  \multicolumn{1}{c|}{/} &
  / &
  \multicolumn{1}{c|}{27009} &
  \multicolumn{1}{c|}{6281} &
  \multicolumn{1}{c|}{17051} &
  \multicolumn{1}{c|}{5666} &
  10573 &
  \multicolumn{1}{c|}{/} &
  \multicolumn{1}{c|}{/} &
  \multicolumn{1}{c|}{/} &
  / \\ \hline
\multicolumn{2}{|c|}{\textbf{AU24}} &
  \multicolumn{1}{c|}{/} &
  \multicolumn{1}{c|}{/} &
  \multicolumn{1}{c|}{/} &
  / &
  \multicolumn{1}{c|}{/} &
  \multicolumn{1}{c|}{/} &
  \multicolumn{1}{c|}{/} &
  \multicolumn{1}{c|}{/} &
  / &
  \multicolumn{1}{c|}{15477} &
  \multicolumn{1}{c|}{3480} &
  \multicolumn{1}{c|}{10108} &
  \multicolumn{1}{c|}{1899} &
  6950 &
  \multicolumn{1}{c|}{/} &
  \multicolumn{1}{c|}{/} &
  \multicolumn{1}{c|}{/} &
  / \\ \hline
\multicolumn{2}{|c|}{\textbf{AU25}} &
  \multicolumn{1}{c|}{92104} &
  \multicolumn{1}{c|}{45461} &
  \multicolumn{1}{c|}{18938} &
  26691 &
  \multicolumn{1}{c|}{11590} &
  \multicolumn{1}{c|}{9473} &
  \multicolumn{1}{c|}{18155} &
  \multicolumn{1}{c|}{3500} &
  6880 &
  \multicolumn{1}{c|}{/} &
  \multicolumn{1}{c|}{/} &
  \multicolumn{1}{c|}{/} &
  \multicolumn{1}{c|}{/} &
  / &
  \multicolumn{1}{c|}{2829} &
  \multicolumn{1}{c|}{1478} &
  \multicolumn{1}{c|}{408} &
  901 \\ \hline
\multicolumn{2}{|c|}{\textbf{AU26}} &
  \multicolumn{1}{c|}{211676} &
  \multicolumn{1}{c|}{26465} &
  \multicolumn{1}{c|}{12532} &
  10440 &
  \multicolumn{1}{c|}{2058} &
  \multicolumn{1}{c|}{1778} &
  \multicolumn{1}{c|}{3146} &
  \multicolumn{1}{c|}{651} &
  1156 &
  \multicolumn{1}{c|}{/} &
  \multicolumn{1}{c|}{/} &
  \multicolumn{1}{c|}{/} &
  \multicolumn{1}{c|}{/} &
  / &
  \multicolumn{1}{c|}{1089} &
  \multicolumn{1}{c|}{578} &
  \multicolumn{1}{c|}{160} &
  334 \\ \hline
\multicolumn{2}{|c|}{\textbf{AU27}} &
  \multicolumn{1}{c|}{/} &
  \multicolumn{1}{c|}{/} &
  \multicolumn{1}{c|}{/} &
  / &
  \multicolumn{1}{c|}{/} &
  \multicolumn{1}{c|}{/} &
  \multicolumn{1}{c|}{/} &
  \multicolumn{1}{c|}{/} &
  / &
  \multicolumn{1}{c|}{/} &
  \multicolumn{1}{c|}{/} &
  \multicolumn{1}{c|}{/} &
  \multicolumn{1}{c|}{/} &
  / &
  \multicolumn{1}{c|}{810} &
  \multicolumn{1}{c|}{382} &
  \multicolumn{1}{c|}{98} &
  313 \\ \hline
\end{tabular}
}
\end{table*}

\begin{table}[]
\centering
\caption{Age Distribution of the Original and New Partitions of GFT }\label{table3}
\setlength{\tabcolsep}{1mm}
\scalebox{0.9}{
\begin{tabular}{|c|c|cccccccc|}
\hline
\multirow{3}{*}{\textbf{Part.}} &
  \multirow{3}{*}{\textbf{Set}} &
  \multicolumn{8}{c|}{\textbf{Age}} \\ \cline{3-10} 
 &
   &
  \multicolumn{1}{c|}{\multirow{2}{*}{\textbf{21}}} &
  \multicolumn{1}{c|}{\multirow{2}{*}{\textbf{22}}} &
  \multicolumn{1}{c|}{\multirow{2}{*}{\textbf{23}}} &
  \multicolumn{1}{c|}{\multirow{2}{*}{\textbf{24}}} &
  \multicolumn{1}{c|}{\multirow{2}{*}{\textbf{25}}} &
  \multicolumn{1}{c|}{\multirow{2}{*}{\textbf{26}}} &
  \multicolumn{1}{c|}{\multirow{2}{*}{\textbf{27}}} &
  \multirow{2}{*}{\textbf{28}} \\
 &
   &
  \multicolumn{1}{c|}{} &
  \multicolumn{1}{c|}{} &
  \multicolumn{1}{c|}{} &
  \multicolumn{1}{c|}{} &
  \multicolumn{1}{c|}{} &
  \multicolumn{1}{c|}{} &
  \multicolumn{1}{c|}{} &
   \\ \hline
\multirow{2}{*}{\textbf{Org.}} &
  Train &
  \multicolumn{1}{c|}{57179} &
  \multicolumn{1}{c|}{18644} &
  \multicolumn{1}{c|}{14004} &
  \multicolumn{1}{c|}{1486} &
  \multicolumn{1}{c|}{3896} &
  \multicolumn{1}{c|}{5797} &
  \multicolumn{1}{c|}{3274} &
  4269 \\ \cline{2-10} 
 &
  Test &
  \multicolumn{1}{c|}{9967} &
  \multicolumn{1}{c|}{10300} &
  \multicolumn{1}{c|}{916} &
  \multicolumn{1}{c|}{1249} &
  \multicolumn{1}{c|}{/} &
  \multicolumn{1}{c|}{/} &
  \multicolumn{1}{c|}{/} &
  2213 \\ \hline
\multirow{3}{*}{\textbf{New}} &
  Train &
  \multicolumn{1}{c|}{35408} &
  \multicolumn{1}{c|}{14918} &
  \multicolumn{1}{c|}{8698} &
  \multicolumn{1}{c|}{1486} &
  \multicolumn{1}{c|}{2237} &
  \multicolumn{1}{c|}{2570} &
  \multicolumn{1}{c|}{1618} &
  2683 \\ \cline{2-10} 
 &
  Valid &
  \multicolumn{1}{c|}{14145} &
  \multicolumn{1}{c|}{4818} &
  \multicolumn{1}{c|}{3096} &
  \multicolumn{1}{c|}{/} &
  \multicolumn{1}{c|}{/} &
  \multicolumn{1}{c|}{1626} &
  \multicolumn{1}{c|}{/} &
  / \\ \cline{2-10} 
 &
  Test &
  \multicolumn{1}{c|}{17593} &
  \multicolumn{1}{c|}{9208} &
  \multicolumn{1}{c|}{3126} &
  \multicolumn{1}{c|}{1249} &
  \multicolumn{1}{c|}{1659} &
  \multicolumn{1}{c|}{1601} &
  \multicolumn{1}{c|}{1656} &
  3799 \\ \hline
\end{tabular}
}
\end{table}

\vspace{-10pt}
\subsection{Performance Measures}

In the following sections, we present the performance measures used to validate the overall performance and fairness of the methods in each task. \\


\noindent
\underline{\textit{\textbf{(1) Expression Recognition}} }
\vspace{0.3cm}
\par

\noindent
\textbf{ I) Global $\text{F}_{1}^{\text{expr}}$:} Evaluates expression recognition performance across the entire partition (training, validation, or test) by averaging the $\text{F}_{1}$ scores of each expression class $c \in C$. It ranges from $[0, 1]$, with higher values being more desirable. It is defined as:
\begin{equation}
\smash{
\text{Global F}_{1}^{\text{expr}} = \frac{1}{|C|} \sum_{c} F_1^c (X)}
\end{equation}
where:  
$X$ denotes the set of all images in the partition (training, validation, or test) and $\text{F}_1^c$ is the $\text{F}_{1}$ Score for the $c$-th expression class across the entire partition. 
%


\noindent
\textbf{II) Local $\text{F}_{1}^{\text{expr}}$:}
Evaluates the fairness performance of expression recognition across all subgroups $g$ of a demographic attribute $G$. It is calculated by averaging the $\text{F}_{1}$ Scores of each subgroup $g$ across all expression classes $c \in C$.  It takes values in $[0, 1]$, with higher values being desirable. It is defined as:
\begin{equation}
\text{Local F}_{1}^{\text{expr}} = \frac{1}{|G| \cdot |C|} \sum_{g} \sum_{c} F_1^c (X_g) 
\end{equation}
where:
$X_g$ denotes the set of all images in the partition annotated with the $g$ subgroup;
$F_1^c (X_g)$ denotes the $\text{F}_{1}$ Score of the  $c$-th expression class, calculated over all images of the $g$-th subgroup.  

\noindent
\textbf{III) Equality of Opportunity (EOP):} 
It ensures that the model performs equally well across different subgroups of a demographic attribute. This prevents the model from favouring one subgroup over another, thus promoting fairness. 
It takes values in $[0,1]$;
values equal or below 0.1 are considered fair.

For each subgroup $g$ of a demographic attribute $G$, at first we compute the confusion matrix $C_g$. We then normalize each of these matrices so that the sum of each row is 1, turning them into error rate matrices. This normalization gives the error rate of predicting class $j$ when the  true class is $i$. A normalised confusion matrix is defined as:
$\hat{C}_g [i,j] = C_g [i,j] \bigl / \sum_j C_g [i,j]  $.

\noindent
Then, we compute the pairwise distances between the normalized confusion matrices using the Mean Absolute Deviation (MAD). For matrices $\hat{C}_{g}$ and $\hat{C}_{g'}$ (of subgroups $g$ and $g'$):
$\text{MAD} (\hat{C}_{g}, \hat{C}_{g'}) =  \sum_{i} \sum_{j} \left | \hat{C}_{g} [i,j] - \hat{C}_{g'} [i,j] \right | \bigl / N^2$. \\
%
%
Finally, we aggregate the pairwise distances to get a single measure of equality of opportunity across all subgroups. We use the mean of these distances. Therefore, EOP is defined as:
\begin{equation}
\text{EOP} = \frac{2}{|G| \cdot (|G| -1)} \sum_{g \in G} \sum_{g' \in G, g \ne g'} \text{MAD} (\hat{C}_{g}, \hat{C}_{g'}).
\end{equation}


\noindent
\textbf{IV) Statistical Parity (SP):} Measures fairness by assessing whether the model predicts outcomes for different demographic subgroups at similar rates. 
It takes values in $[0, 1]$
; values equal or below 0.1 are considered fair.
At first, for each subgroup $g \in G$ and each expression category $c \in C$ we compute the success rate:  
$\text{SR}_g^c = \sum_{i \in G_k} \mathds{1} (\hat{y}_i = c) \bigl / |G_k| $.
\hspace{0.02cm} $G_k$ is the set of indices of samples belonging to subgroup $g$; $\hat{y_i}$ is the predicted class for sample $i$ and $ \mathds{1}(\cdot)$ is the indicator function.
Then, for each group $g$ we construct the success rate vector $\text{SR}_g$ of length  $|C|: \text{SR}_g = [\text{SR}_g^{c_1}, \text{SR}_g^{c_2}, ..., \text{SR}_g^C]$. Then, for each pair of groups $(g, g')$, we compute the MAD between their success rate vectors:
$\text{MAD} (\text{SR}_g, \text{SR}_{g'}) =  \sum_{c \in C} \left |\text{SR}_g^c - \text{SR}_{g'}^c \right | \bigl / |C| $.  
\hspace{0.02cm}Finally, we average across all pairwise MADs  and thus SP is defined as:
\begin{equation}
\text{SP} = \frac{2}{|G| \cdot (|G| -1)} \sum_{g \in G} \sum_{g' \in G, g \ne g'} \text{MAD} (\text{SR}_g, \text{SR}_{g'}).
\end{equation}


\vspace{0.2em}
\noindent
\underline{\textit{\textbf{(2) Action Unit Detection}} }
\vspace{0.2cm}

\noindent
\textbf{I) Global $\text{F}_{1}^{\text{AU}}$:} Evaluates the overall performance of AU Detection across the entire partition (i.e., training, validation, or test set). It is calculated -similarly to $\text{Global F}_{1}^{\text{expr}}$- by averaging the $\text{F}_{1}$ scores of each AU over the whole partition. It takes values in $[0, 1]$, with higher values being desirable. 



\noindent
\textbf{II) Local $\text{F}_{1}^{\text{AU}}$:} Evaluates the fairness performance of AU detection across all subgroups $g$ of a demographic attribute $G$.  It takes values in $[0, 1]$, with higher values being desirable.
It is calculated -similarly to $\text{Local F}_{1}^{\text{expr}}$- by averaging the F1 Scores of each subgroup $g$ across all AUs. 


\noindent
\textbf{III) Equal Opportunity Difference (EOD):} 
It measures fairness by ensuring that the True Positive Rates (TPR) for each AU are similar across different demographic subgroups. This helps to avoid biases where some subgroups might have significantly higher or lower TPRs for certain AUs, leading to unfair advantages or disadvantages.
It takes values in $[0, 1]$; values equal or below 0.1 are considered fair.
For each subgroup $g \in G$ and each $au$, at first, we compute the: 
$\text{TPR}_{g}^{\text{au}} = 
  \sum_{i \in G_k} \mathds{1} \left ( \hat{y}_i^{au} = 1 \land y_i^{au} = 1 \right ) \bigl /  \sum_{i \in G_k} \mathds{1} \left ( y_i^{au} = 1 \right )  $.
\hspace{0.02cm} $G_k$ is the set of indices of samples belonging to subgroup $g$; $\hat{y}_i^{au}$ is the prediction for the specific $au$ for sample $i$; $y_i^{au}$ is the true label for the specific $au$ for sample $i$. 
Then, for each $au$, we calculate the difference between the maximum and minimum TPRs across all subgroups; finally, EOD is the average of these differences across all $M$ AUs:
\begin{equation}
\text{EOD} = \frac{1}{M} \sum_{au} \left ( \max_{g \in G} \text{TPR}_g^{au} - \min_{g \in G} \text{TPR}_g^{au} \right ) .
\end{equation}


\noindent
\textbf{IV) Demographic Parity Difference (DPD):} 
It measures fairness by comparing the selection rates across different demographic subgroups. It is defined as the difference between the largest and smallest group-level selection rates. It takes values in $[0, 1]$
; values equal or below 0.1 are considered fair.
At first, for each subgroup $g \in G$ we compute the selection rate for each $au$:
$\text{SR}_g^{au} = \sum_{i \in G_k} \mathds{1} (\hat{y}_i^{au} = 1) \bigl / |G_k|$.
\hspace{0.02cm}
$G_k$ is the set of indices of samples belonging to subgroup $g$; $\hat{y}_i^{au}$ is the predicted AU for sample $i$ and $ \mathds{1}(\cdot)$ is the indicator function.
Then, for each $au$, we calculate the difference between the maximum and minimum selection rates across all subgroups $g$; DPD is the average of these differences across all $M$ AUs:
\begin{equation}
\text{DPD} = \frac{1}{M} \sum_{au} \left ( \max_{g \in G} \text{SR}_g^{au} - \min_{g \in G} \text{SR}_g^{au} \right ) .
\end{equation}

\vspace{0.2em}
\noindent
\underline{\textit{\textbf{(3) Valence-Arousal Estimation}} }
\vspace{0.5em}
\par

\noindent
\textbf{I) Global CCC:} 
Evaluates the overall performance of VA Estimation across the entire partition (i.e., training, validation, or test
set). It takes values in $[-1, 1]$, with higher values being desirable. It is calculated as:
\begin{equation}
\text{Global CCC}^{\text{VA}} = \frac{1}{2} \left ( \text{CCC}_X^{\text{V}} + \text{CCC}_X^{\text{A}}  \right )
\end{equation}
where: $ \text{CCC}_X^i = 2 \cdot s_{xy} / \left[ s_x^2 + s_y^2 + (\bar{x} - \bar{y})^2 \right]$, with $i =$ V or A, and $X$ being the set of all images in the partition; $s_x$ and $s_y$ represent the variances of the valence/arousal annotations and predicted values, respectively; $\bar{x}$ and $\bar{y}$ denote the mean values of the corresponding annotations and predictions; $s_{xy}$ represents their covariance. 

\noindent
\textbf{II) Local CCC:} Evaluates the alignment between predicted values and annotations across diverse demographic subgroups within a demographic attribute. It is computed by averaging the mean CCC values across all subgroups:
\begin{equation}
\smash{
\text{Local CCC}^{\text{VA}} = \frac{1}{2 \cdot |G|} \sum_g  \left ( \text{CCC}_{X_g}^{\text{V}} + \text{CCC}_{X_g}^{\text{A}}  \right )}
\end{equation}
where: $X_g$ is the set of all images in the partition
annotated with the $g$ subgroup; $\text{CCC}_{X_g}^i$ denotes the CCC
of valence/arousal calculated over all images of the $g$-th subgroup.


\section{EXPERIMENTAL RESULTS}

In the following, we provide an extensive experimental study in which we utilise the most widely used baseline models (variants of ResNet, VGG, DenseNet, ConvNext, EfficientNet, Swin and ViT), the top-performing methods from the ABAW Competitions 
(FUXI \cite{fuxi}, SITU \cite{situ}, CTC \cite{ctc}), and the state-of-the-art (SOTA) methods (EmoGCN \cite{emogcn}, MT-EffNet \cite{mteffnet}, DACL \cite{dacl}, MA-Net \cite{manet} \& EAC \cite{eac}, DAN \cite{dan}, POSTER++ \cite{poster}; ME-GraphAU \cite{megraphau} \& AUNets \cite{aunets})
for Expression Recognition, AU Detection and VA Estimation. We train these models on each examined database -split with the proposed partitioning protocol- and evaluate them in terms of the metrics defined in the previous Section, both the overall-global metrics, as well as the metrics that measure the fairness of the methods.


\vspace{-10pt}

\subsection{Expression Recognition Performance}

\begin{table}[]
\centering
\caption{Performance (in \%) of various baseline and SOTA methods for Expression Recognition. For $\text{GF}_{1}$ and $\text{LF}_{1}$  higher values are wanted; for SP and EOP values in $[0,10]$ indicate fair methods. GF$_1$ stands for Global $F_1^{expr}$; LF$_1$ stands for Local $F_1^{expr}$. 
}\label{table4}
\setlength{\tabcolsep}{1mm}
\renewcommand{\arraystretch}{1}
\scalebox{.85}{
\begin{tabular}{|c|cccccccccc|}
\hline
\multirow{3}{*}{\textbf{Model}} & \multicolumn{10}{c|}{\textbf{RAF-DB (test set)}}                                                                                                                                                                                                                                                                                                      \\ \cline{2-11} 
                                & \multicolumn{1}{c|}{\textbf{Whole}} & \multicolumn{3}{c|}{\textbf{Race}}                                                                         & \multicolumn{3}{c|}{\textbf{Gender}}                                                                       & \multicolumn{3}{c|}{\textbf{Age}}                                                     \\ \cline{2-11} 
                                & \multicolumn{1}{c|}{\textbf{GF$_1$}} & \multicolumn{1}{c|}{\textbf{SP}} & \multicolumn{1}{c|}{\textbf{LF$_1$}} & \multicolumn{1}{c|}{\textbf{EOP}} & \multicolumn{1}{c|}{\textbf{SP}} & \multicolumn{1}{c|}{\textbf{LF$_1$}} & \multicolumn{1}{c|}{\textbf{EOP}} & \multicolumn{1}{c|}{\textbf{SP}} & \multicolumn{1}{c|}{\textbf{LF$_1$}} & \textbf{EOP} \\ \hline \hline
\textbf{ResNet18}               & \multicolumn{1}{c|}{64.1}           & \multicolumn{1}{c|}{6.4}         & \multicolumn{1}{c|}{63.0}           & \multicolumn{1}{c|}{13.0}         & \multicolumn{1}{c|}{9.9}         & \multicolumn{1}{c|}{63.2}           & \multicolumn{1}{c|}{10.0}         & \multicolumn{1}{c|}{24.3}        & \multicolumn{1}{c|}{57.1}           & 27.5         \\ \hline
\textbf{ResNet34}               & \multicolumn{1}{c|}{65.6}           & \multicolumn{1}{c|}{7.5}         & \multicolumn{1}{c|}{64.2}           & \multicolumn{1}{c|}{13.5}         & \multicolumn{1}{c|}{9.1}         & \multicolumn{1}{c|}{64.7}           & \multicolumn{1}{c|}{10.4}         & \multicolumn{1}{c|}{23.0}        & \multicolumn{1}{c|}{60.3}           & 25.0         \\
\hline
\end{tabular}
}
\end{table}

\begin{table*}[]
\centering
\caption{Performance (in \%) of various baseline and state-of-the-art methods for Expression Recognition. For $\text{GF}_{1}$ and $\text{LF}_{1}$  higher values are wanted; for SP and EOP values in $[0,10]$ indicate fair methods. GF$_1$ stands for Global $F_1^{expr}$; LF$_1$ stands for Local $F_1^{expr}$
}\label{table5}
\renewcommand{\arraystretch}{1}
\scalebox{.8}{
\begin{tabular}{|c|cccccccccc|cccccccccc|}
\hline
\multirow{3}{*}{\textbf{Model}} & \multicolumn{10}{c|}{\textbf{AffectNet-7 (test set)}}                                                                                                                                                                                                                                                                                                         & \multicolumn{10}{c|}{\textbf{AffectNet-8 (test set)}}                                                                                                                                                                                                                                                                                                         \\ \cline{2-21} 
                                & \multicolumn{1}{c|}{\textbf{Whole}}   & \multicolumn{3}{c|}{\textbf{Race}}                                                                           & \multicolumn{3}{c|}{\textbf{Gender}}                                                                         & \multicolumn{3}{c|}{\textbf{Age}}                                                       & \multicolumn{1}{c|}{\textbf{Whole}}   & \multicolumn{3}{c|}{\textbf{Race}}                                                                           & \multicolumn{3}{c|}{\textbf{Gender}}                                                                         & \multicolumn{3}{c|}{\textbf{Age}}                                                       \\
\hline
\end{tabular}
}
\end{table*}

Tables \ref{table4} and \ref{table5} present the Global F$_1^{expr}$ performance of these methods on RAF-DB and AffectNet, respectively. In Table \ref{table4}, ConvNeXt\_l (top-performing among ConvNeXt variants) surpasses Swin\_b (top-performing among Swin variants), which outperforms EfficientNet\_v2\_l and DenseNet161, these results are expected. It is also expected that the largest models of the same variants yield the best performance on RAF-DB which is a relatively small database (e.g., ConvNeXt\_l outperforms ConvNeXt\_b, which outperforms ConvNeXt\_t). However, surprisingly, VGG11 achieves better performance than iResNet101 (which outperforms ResNext101\_64 and ResNet152). Additionally, Swin Transformer variants outperform ViT variants (the same stood in the case of both AffectNet-7 and AffectNet-8 from Table \ref{table5}); the former producing higher performance than all baseline models except ConvNeXt\_l, which shows a significant performance difference of more than 2\%.

For the ABAW models, CTC now outperforms SITU on the new partition of AffectNet-8, while SITU had previously outperformed CTC on the original partition. Additionally, DACL, which had the worst performance among all SOTA models on the old RAF-DB partition, now shows the second-best performance after the highest performance POSTER++ on both RAF-DB and AffectNet-7. Notably, many baseline models achieve higher performance than several SOTA models on both RAF-DB and AffectNet-7.

Table \ref{table5} illustrates that on AffectNet-7, ConvNeXt\_t achieve the best performance among all baseline methods. DenseNet201 the second performance model, performs better than EfficientNet\_v2\_m, ResNeXt101\_64, ResNet152, and iResNet101. VGG variants achieve the worst performance on this large-scale database (both for AffectNet-7 and -8). In terms of SOTA methods, it is important to note that in this new partition, DACL has the same performance as EAC and both outperform DAN, whereas in the original partition, DAN outperformed EAC, which outperformed DACL. As we mentioned, such changes in performance ranking among SOTA methods (with the new partition protocol) in all examined databases, underscore the importance of a robust and large test set in accurately assessing model performance. These observed changes suggest that the previous partitions may have been insufficiently representative evaluations. By using the proposed protocol, we enhance the credibility and robustness of the database evaluation process.

Table \ref{table5} illustrates the performance on AffectNet-8, where FUXI achieves the best overall performance, surpassing CTC, which outperforms SITU. In the original partition, SITU outperformed CTC, indicating a shift in performance rankings. Notably, among the SOTA methods in the new partition, EmoGCN ranks as the second-best performing model (previously fifth in the original partition), followed by DACL as the third (previously the worst performing), MT-EffNet as the fourth (previously second), EAC as the fifth (previously sixth), DAN as the sixth (previously third), and MA-Net as the lowest performing SOTA (previously fourth). Therefore we observe a completely different ranking of these methods. 
Finally, Table \ref{table5} shows that EfficientNet\_v2\_s outperforms DenseNet161, which outperforms ConvNeXt\_t, which outperforms ResNext101\_64 and iResNet101, which outperform ResNet34.


Tables \ref{table4} and \ref{table5} further illustrate that w.r.t. to SP metric: 
i) all methods are fair for race in all databases (RAF-DB and AffectNet), since all scores are less than 10; 
ii) all methods are almost fair for gender on RAF-DB, since all scores are around 10; all methods are not considered fair for gender on AffectNet (both versions) but the achieved performance is quite close to count as fair as the scores are around 17;
iii) all methods are not fair for age in all databases; in RAF-DB the scores are more than 20, whereas in AffectNet (both versions) the scores are lower, around 15, which are close to be considered fair.

Tables \ref{table4} and \ref{table5} further illustrate that w.r.t. to EOP metric: 
i) all methods are fair for race in both versions of AffectNet (with values below 10) and all methods are  not fair but close to be considered fair in RAF-DB (with values around 14);
ii) all methods are fair for gender in both versions of AffectNet (with values below 10) and all methods are either fair or very close to be considered fair in RAF-DB (with values between 6 and 12); it is worth to mention that POSTER++ was the best performing method on RAF-DB and achieved the lowest EOP score for gender making it also the fairest method;
iii) all methods are not fair for age in RAF-DB with values around 24; all methods are not considered fair for age in both versions of AffectNet but are close to be considered fair (since values are around 15); it is worth noting again that POSTER++ achieved the lowest EOP score for age in RAF-DB.

Finally, Tables \ref{table4} and \ref{table5}  illustrate that w.r.t. to Local F$_1^{expr}$, POSTER++ is the top-performing method (which is consistent with our previous observations) and all scores for race and gender were close in value to Global F$_1^{expr}$; for age the scores were a bit lower than the Global F$_1^{expr}$.

All the above findings challenge the conventional notion that achieving fairness inevitably leads to a decrease in overall performance. On the contrary, our findings suggest that models can achieve both fairness and better overall performance concurrently. This indicates that there is not necessarily an inherent trade-off between fairness and performance, debunking a common conception of fair model design. This insight provides valuable guidance for future fair model development, emphasising the importance of prioritising fairness without compromising performance.

\vspace{-10pt}
\subsection{AU Detection Performance}

\begin{table*}[]
\scriptsize
\centering
\caption{
Performance (in \%) of various baseline and state-of-the-art methods for AU Detection. For $\text{GF}_{1}$ and $\text{LF}_{1}$  higher values are wanted; for EOD and DPD values in $[0,10]$ indicate fair methods. GF$_1$ stands for Global $F_1^{expr}$; LF$_1$ stands for Local $F_1^{expr}$. 
}\label{table6}
\setlength{\tabcolsep}{2.2mm}
\renewcommand{\arraystretch}{1}
\scalebox{.8}{
\begin{tabular}{|c|cccccccccc|cccccccccc|}
\hline
\multirow{3}{*}{\textbf{Model}} & \multicolumn{10}{c|}{\textbf{RAF-AU (test set)}}                                                                                                                                                                                                                                                                                                                        & \multicolumn{10}{c|}{\textbf{EmotioNet (test set)}}                                                                                                                                                                                                                                                                                                                     \\ \cline{2-21} 
                                & \multicolumn{1}{c|}{\textbf{Whole}}  & \multicolumn{3}{c|}{\textbf{Race}}                                                                           & \multicolumn{3}{c|}{\textbf{Gender}}                                                                         & \multicolumn{3}{c|}{\textbf{Age}}                                                       & \multicolumn{1}{c|}{\textbf{Whole}}  & \multicolumn{3}{c|}{\textbf{Race}}                                                                           & \multicolumn{3}{c|}{\textbf{Gender}}                                                                         & \multicolumn{3}{c|}{\textbf{Age}}                                                       \\
\hline
\end{tabular}
}
\end{table*}

Table \ref{table6} illustrates the overall and fairness performance of all models on the new partitions of RAF-AU, EmotioNet, DISFA, and GFT dataset. In terms of Global F$_1^{au}$, ConvNeXt variants consistently outperform other models across DISFA, RAF-AU, and GFT, followed by DenseNet and VGG variants. On DISFA, the ResNeXt variant follows, outperforming EfficientNet, ResNet, and iResNet variants. However, on RAF-AU and EmotioNet, EfficientNet, ResNet, and iResNet variants follow, with ResNeXt demonstrating the lowest performance.
Interestingly, on GFT, EffNet\_b6 (the top-performing EfficientNet variant) achieves the best performance among all baseline methods, surpassing ResNet50 and iResNet101 (the worst models on DISFA). ConvNeXt, DenseNet, and VGG11, which are the top three on other datasets, rank third, fourth, and last among baseline models on GFT.
Additionally, Swin Transformer variants consistently outperform ViT variants across all datasets, similar to the FER task. On RAF-AU and GFT, their performance variance is small (around 1\%), but on EmotioNet and DISFA, significant differences are observed, with a remarkable 6.5\% difference on EmotioNet.

Notably, the ABAW models (CTC, FUXI, and SITU) exhibit diverse performance across the four datasets. In the old partition, FUXI outperformed SITU, which outperformed CTC. In the new partition, CTC performs best on EmotioNet, GFT, and DISFA but worst on RAF-AU. SITU outperforms FUXI on RAF-AU and GFT, while for EmotioNet and DISFA are the opposite. Among the SOTA methods, AUNets consistently outperform ME-GraphAU across all datasets, with a significant performance gap. On EmotioNet, this gap is nearly 10\%, compared to an average difference of 1.7\% in the old partition. This highlights the improved robustness and generalization of AUNets in the new partition settings.

Table \ref{table6} illustrates the fairness performance of all methods with respect to the EOD metric:
i) None of the methods are fair regarding race across all databases, with EOD values around 20 on RAF-DB and EmotioNet, around 30 on GFT, and around 50 on DISFA. The worst model, EffNet\_b2, reaches an EOD value of 60.5.
ii) For the gender attribute, EOD values on RAF-AU and EmotioNet indicate fairness, with scores below 10. On DISFA and GFT, EOD values are around 15, which is close to being considered fair.
iii) None of the methods are fair for age across all databases, with EOD values around 50 on RAF-AU and DISFA, around 35 on EmotioNet, and exceeding 50 on GFT.
iv) The EOD metric effectively reveals demographic attribute distributions within the datasets. It shows fair performance on balanced attributes (e.g., gender) and large biases on skewed attributes (e.g., age).


Table \ref{table6} further illustrates the fairness performance for the DPD metric:
i) All methods are fair concerning race in EmotioNet, with scores below 5. In RAF-AU, DPD scores are around 15, indicating near fairness. However, on the DISFA dataset, scores around 20 indicate unfairness.
ii) For the gender attribute, all methods are fair across all databases. DPD scores are under 10 for RAF-AU, EmotioNet, and DISFA, and close to 10 for GFT.
iii) Regarding the age attribute, only EmotioNet shows near fairness, with scores under 15. Scores for other databases exceed 20, indicating unfairness.
iv) Notably, models with the worst Global F$_1^{au}$ performance, such as ResNet50\_32 on RAF-AU, ViT\_l\_32 on EmotioNet, EffNet\_b0 on DISFA, and ME-GraphAU on EmotioNet, exhibit relatively low DPD scores. This suggests these models may have lower overall detection rates, reducing disparities in prediction rates across demographic groups.
v) The best Global F$_1^{au}$ model, AUNets on RAF-AU and EmotioNet, also shows relatively low DPD scores across age, gender, and race. However, the best-performing model, CTC, on DISFA and GFT exhibits higher bias compared to other methods.

Table \ref{table6} illustrates Local F$_1^{au}$ scores. The scores for gender closely match the Global F$_1^{au}$, indicating consistent performance across gender groups. For race attribute, Local F$_1^{au}$ scores are slightly lower than Global F$_1^{au}$, suggesting minor disparities in performance across different race groups. The most significant difference is observed in age, with Local F$_1^{au}$ scores approximately 5\% lower than Global F$_1^{au}$, indicating a pronounced bias in performance across age groups.

\vspace{-10pt}

\subsection{VA Estimation}

\begin{table}[]
\centering
\caption{CCC Performance comparison (in \%) between various baseline and state-of-the-art works for VA estimation, A higher value is desired. GVA stands for Global CCC$^{\text{VA}}$; GV stands for Global CCC$^{\text{V}}$; GA stands for Global CCC$^{\text{A}}$; LVA stands for Local CCC$^{\text{VA}}$; LV stands for Local CCC$^{\text{V}}$; LA stands for Local CCC$^{\text{A}}$;}
\label{table7}
\setlength{\tabcolsep}{1mm}
\renewcommand{\arraystretch}{1.3}
\resizebox{\columnwidth}{!}{
\begin{tabular}{|c|cccccccccccc|}
\hline
\multirow{3}{*}{\textbf{Model}} & \multicolumn{12}{c|}{\textbf{AffectNet-VA}}                                                                                                                                                                                                                                                                                                                                                                      \\ \cline{2-13} 
                                & \multicolumn{3}{c|}{\textbf{Whole}}                                                                     & \multicolumn{3}{c|}{\textbf{Race}}                                                                      & \multicolumn{3}{c|}{\textbf{Gender}}                                                                    & \multicolumn{3}{c|}{\textbf{Age}}                                                  \\ \cline{2-13} 
                                & \multicolumn{1}{c|}{\textbf{GVA}} & \multicolumn{1}{c|}{\textbf{GV}} & \multicolumn{1}{c|}{\textbf{GA}} & \multicolumn{1}{c|}{\textbf{LVA}} & \multicolumn{1}{c|}{\textbf{LV}} & \multicolumn{1}{c|}{\textbf{LA}} & \multicolumn{1}{c|}{\textbf{LVA}} & \multicolumn{1}{c|}{\textbf{LV}} & \multicolumn{1}{c|}{\textbf{LA}} & \multicolumn{1}{c|}{\textbf{LVA}} & \multicolumn{1}{c|}{\textbf{LV}} & \textbf{LA} \\ \hline  \hline
\textbf{ResNet18}               & \multicolumn{1}{c|}{69.8}         & \multicolumn{1}{c|}{81.2}        & \multicolumn{1}{c|}{58.4}        & \multicolumn{1}{c|}{69.2}         & \multicolumn{1}{c|}{81.5}        & \multicolumn{1}{c|}{56.8}        & \multicolumn{1}{c|}{69.3}         & \multicolumn{1}{c|}{80.4}        & \multicolumn{1}{c|}{58.1}        & \multicolumn{1}{c|}{68.8}         & \multicolumn{1}{c|}{80.7}        & 57.0        \\ \hline
\textbf{ResNet34}               & \multicolumn{1}{c|}{70.5}         & \multicolumn{1}{c|}{81.9}        & \multicolumn{1}{c|}{59.8}        & \multicolumn{1}{c|}{69.8}         & \multicolumn{1}{c|}{82.2}        & \multicolumn{1}{c|}{57.4}        & \multicolumn{1}{c|}{70.0}         & \multicolumn{1}{c|}{81.2}        & \multicolumn{1}{c|}{58.8}        & \multicolumn{1}{c|}{69.6}         & \multicolumn{1}{c|}{81.4}        & 57.7        \\ \hline
\textbf{ResNet50}               & \multicolumn{1}{c|}{68.9}         & \multicolumn{1}{c|}{80.6}        & \multicolumn{1}{c|}{57.3}        & \multicolumn{1}{c|}{67.9}         & \multicolumn{1}{c|}{80.7}        & \multicolumn{1}{c|}{55.1}        & \multicolumn{1}{c|}{68.4}         & \multicolumn{1}{c|}{79.8}        & \multicolumn{1}{c|}{57.0}        & \multicolumn{1}{c|}{67.5}         & \multicolumn{1}{c|}{79.9}        & 55.1        \\ \hline
\textbf{ResNet101}              & \multicolumn{1}{c|}{70.4}         & \multicolumn{1}{c|}{81.6}        & \multicolumn{1}{c|}{59.3}        & \multicolumn{1}{c|}{69.8}         & \multicolumn{1}{c|}{81.8}        & \multicolumn{1}{c|}{57.8}        & \multicolumn{1}{c|}{70.0}         & \multicolumn{1}{c|}{80.9}        & \multicolumn{1}{c|}{59.1}        & \multicolumn{1}{c|}{69.0}         & \multicolumn{1}{c|}{80.8}        & 57.2        \\ \hline
\textbf{ResNet152}              & \multicolumn{1}{c|}{70.1}         & \multicolumn{1}{c|}{81.3}        & \multicolumn{1}{c|}{59.3}        & \multicolumn{1}{c|}{69.3}         & \multicolumn{1}{c|}{81.5}        & \multicolumn{1}{c|}{57.2}        & \multicolumn{1}{c|}{69.7}         & \multicolumn{1}{c|}{80.6}        & \multicolumn{1}{c|}{58.8}        & \multicolumn{1}{c|}{69.0}         & \multicolumn{1}{c|}{80.7}        & 57.3        \\ \hline  \hline
\textbf{ResNext50\_32}          & \multicolumn{1}{c|}{68.9}         & \multicolumn{1}{c|}{79.9}        & \multicolumn{1}{c|}{58.1}        & \multicolumn{1}{c|}{68.1}         & \multicolumn{1}{c|}{79.9}        & \multicolumn{1}{c|}{56.4}        & \multicolumn{1}{c|}{68.3}         & \multicolumn{1}{c|}{78.8}        & \multicolumn{1}{c|}{57.8}        & \multicolumn{1}{c|}{67.5}         & \multicolumn{1}{c|}{79.1}        & 55.8        \\ \hline
\textbf{ResNext101\_32}         & \multicolumn{1}{c|}{70.4}         & \multicolumn{1}{c|}{80.8}        & \multicolumn{1}{c|}{59.9}        & \multicolumn{1}{c|}{70.0}         & \multicolumn{1}{c|}{81.3}        & \multicolumn{1}{c|}{58.6}        & \multicolumn{1}{c|}{69.8}         & \multicolumn{1}{c|}{80.1}        & \multicolumn{1}{c|}{59.6}        & \multicolumn{1}{c|}{69.1}         & \multicolumn{1}{c|}{80.4}        & 57.7        \\ \hline
\textbf{ResNext101\_64}         & \multicolumn{1}{c|}{70.8}         & \multicolumn{1}{c|}{81.5}        & \multicolumn{1}{c|}{60.1}        & \multicolumn{1}{c|}{70.3}         & \multicolumn{1}{c|}{81.8}        & \multicolumn{1}{c|}{58.8}        & \multicolumn{1}{c|}{70.2}         & \multicolumn{1}{c|}{80.7}        & \multicolumn{1}{c|}{59.8}        & \multicolumn{1}{c|}{69.5}         & \multicolumn{1}{c|}{81.0}        & 57.9        \\ \hline  \hline
\textbf{iResNet101}             & \multicolumn{1}{c|}{71.0}         & \multicolumn{1}{c|}{82.1}        & \multicolumn{1}{c|}{60.3}        & \multicolumn{1}{c|}{70.3}         & \multicolumn{1}{c|}{82.1}        & \multicolumn{1}{c|}{58.6}        & \multicolumn{1}{c|}{70.5}         & \multicolumn{1}{c|}{81.0}        & \multicolumn{1}{c|}{60.1}        & \multicolumn{1}{c|}{69.9}         & \multicolumn{1}{c|}{81.2}        & 58.5        \\ \hline  \hline
\textbf{VGG11}                  & \multicolumn{1}{c|}{69.7}         & \multicolumn{1}{c|}{80.9}        & \multicolumn{1}{c|}{58.6}        & \multicolumn{1}{c|}{68.9}         & \multicolumn{1}{c|}{81.2}        & \multicolumn{1}{c|}{56.5}        & \multicolumn{1}{c|}{69.3}         & \multicolumn{1}{c|}{80.1}        & \multicolumn{1}{c|}{58.4}        & \multicolumn{1}{c|}{68.7}         & \multicolumn{1}{c|}{80.3}        & 57.1        \\ \hline
\textbf{VGG16}                  & \multicolumn{1}{c|}{70.7}         & \multicolumn{1}{c|}{81.4}        & \multicolumn{1}{c|}{60.5}        & \multicolumn{1}{c|}{70.2}         & \multicolumn{1}{c|}{81.3}        & \multicolumn{1}{c|}{59.1}        & \multicolumn{1}{c|}{70.2}         & \multicolumn{1}{c|}{80.3}        & \multicolumn{1}{c|}{60.2}        & \multicolumn{1}{c|}{69.5}         & \multicolumn{1}{c|}{80.4}        & 58.7        \\ \hline
\textbf{VGG19}                  & \multicolumn{1}{c|}{71.0}         & \multicolumn{1}{c|}{81.4}        & \multicolumn{1}{c|}{60.7}        & \multicolumn{1}{c|}{70.4}         & \multicolumn{1}{c|}{81.6}        & \multicolumn{1}{c|}{59.3}        & \multicolumn{1}{c|}{70.5}         & \multicolumn{1}{c|}{80.6}        & \multicolumn{1}{c|}{60.4}        & \multicolumn{1}{c|}{69.8}         & \multicolumn{1}{c|}{80.7}        & 58.8        \\ \hline  \hline
\textbf{DenseNet121}            & \multicolumn{1}{c|}{69.8}         & \multicolumn{1}{c|}{81.5}        & \multicolumn{1}{c|}{58.1}        & \multicolumn{1}{c|}{69.3}         & \multicolumn{1}{c|}{81.8}        & \multicolumn{1}{c|}{56.9}        & \multicolumn{1}{c|}{69.4}         & \multicolumn{1}{c|}{80.8}        & \multicolumn{1}{c|}{57.9}        & \multicolumn{1}{c|}{68.5}         & \multicolumn{1}{c|}{80.9}        & 56.2        \\ \hline
\textbf{DenseNet161}            & \multicolumn{1}{c|}{70.6}         & \multicolumn{1}{c|}{81.7}        & \multicolumn{1}{c|}{60.6}        & \multicolumn{1}{c|}{69.8}         & \multicolumn{1}{c|}{82.0}        & \multicolumn{1}{c|}{57.6}        & \multicolumn{1}{c|}{70.2}         & \multicolumn{1}{c|}{81.1}        & \multicolumn{1}{c|}{59.3}        & \multicolumn{1}{c|}{69.5}         & \multicolumn{1}{c|}{81.2}        & 57.9        \\ \hline
\textbf{DenseNet201}            & \multicolumn{1}{c|}{70.2}         & \multicolumn{1}{c|}{81.9}        & \multicolumn{1}{c|}{58.9}        & \multicolumn{1}{c|}{69.7}         & \multicolumn{1}{c|}{82.4}        & \multicolumn{1}{c|}{57.1}        & \multicolumn{1}{c|}{69.7}         & \multicolumn{1}{c|}{81.2}        & \multicolumn{1}{c|}{58.3}        & \multicolumn{1}{c|}{69.2}         & \multicolumn{1}{c|}{81.4}        & 57.0        \\ \hline  \hline
\textbf{ConvNeXt\_t}            & \multicolumn{1}{c|}{72.0}         & \multicolumn{1}{c|}{82.7}        & \multicolumn{1}{c|}{61.2}        & \multicolumn{1}{c|}{71.3}         & \multicolumn{1}{c|}{83.0}        & \multicolumn{1}{c|}{59.7}        & \multicolumn{1}{c|}{71.6}         & \multicolumn{1}{c|}{82.2}        & \multicolumn{1}{c|}{61.0}        & \multicolumn{1}{c|}{71.0}         & \multicolumn{1}{c|}{82.4}        & 59.7        \\ \hline
\textbf{ConvNeXt\_s}            & \multicolumn{1}{c|}{71.9}         & \multicolumn{1}{c|}{82.6}        & \multicolumn{1}{c|}{61.5}        & \multicolumn{1}{c|}{71.4}         & \multicolumn{1}{c|}{82.7}        & \multicolumn{1}{c|}{60.2}        & \multicolumn{1}{c|}{71.5}         & \multicolumn{1}{c|}{81.8}        & \multicolumn{1}{c|}{61.2}        & \multicolumn{1}{c|}{71.0}         & \multicolumn{1}{c|}{82.1}        & 59.9        \\ \hline
\textbf{ConvNeXt\_b}            & \multicolumn{1}{c|}{72.4}         & \multicolumn{1}{c|}{82.9}        & \multicolumn{1}{c|}{61.9}        & \multicolumn{1}{c|}{71.9}         & \multicolumn{1}{c|}{83.3}        & \multicolumn{1}{c|}{60.6}        & \multicolumn{1}{c|}{72.0}         & \multicolumn{1}{c|}{82.3}        & \multicolumn{1}{c|}{61.7}        & \multicolumn{1}{c|}{71.5}         & \multicolumn{1}{c|}{82.5}        & 60.5        \\ \hline
\textbf{ConvNeXt\_l}            & \multicolumn{1}{c|}{72.3}         & \multicolumn{1}{c|}{82.9}        & \multicolumn{1}{c|}{61.7}        & \multicolumn{1}{c|}{71.8}         & \multicolumn{1}{c|}{83.2}        & \multicolumn{1}{c|}{60.5}        & \multicolumn{1}{c|}{71.8}         & \multicolumn{1}{c|}{82.3}        & \multicolumn{1}{c|}{61.4}        & \multicolumn{1}{c|}{71.4}         & \multicolumn{1}{c|}{82.5}        & 60.2        \\ \hline  \hline
\textbf{EffNet\_b0}             & \multicolumn{1}{c|}{71.2}         & \multicolumn{1}{c|}{82.0}        & \multicolumn{1}{c|}{60.3}        & \multicolumn{1}{c|}{70.5}         & \multicolumn{1}{c|}{82.3}        & \multicolumn{1}{c|}{58.7}        & \multicolumn{1}{c|}{70.7}         & \multicolumn{1}{c|}{81.3}        & \multicolumn{1}{c|}{60.0}        & \multicolumn{1}{c|}{70.2}         & \multicolumn{1}{c|}{81.6}        & 58.7        \\ \hline
\textbf{EffNet\_b1}             & \multicolumn{1}{c|}{71.0}         & \multicolumn{1}{c|}{82.0}        & \multicolumn{1}{c|}{60.1}        & \multicolumn{1}{c|}{70.4}         & \multicolumn{1}{c|}{82.1}        & \multicolumn{1}{c|}{58.7}        & \multicolumn{1}{c|}{70.5}         & \multicolumn{1}{c|}{81.2}        & \multicolumn{1}{c|}{59.8}        & \multicolumn{1}{c|}{69.9}         & \multicolumn{1}{c|}{81.4}        & 58.4        \\ \hline
\textbf{EffNet\_b2}             & \multicolumn{1}{c|}{71.1}         & \multicolumn{1}{c|}{82.4}        & \multicolumn{1}{c|}{59.8}        & \multicolumn{1}{c|}{70.5}         & \multicolumn{1}{c|}{82.7}        & \multicolumn{1}{c|}{58.4}        & \multicolumn{1}{c|}{70.6}         & \multicolumn{1}{c|}{81.7}        & \multicolumn{1}{c|}{59.4}        & \multicolumn{1}{c|}{69.9}         & \multicolumn{1}{c|}{81.8}        & 57.9        \\ \hline
\textbf{EffNet\_b6}             & \multicolumn{1}{c|}{71.6}         & \multicolumn{1}{c|}{82.3}        & \multicolumn{1}{c|}{60.8}        & \multicolumn{1}{c|}{70.9}         & \multicolumn{1}{c|}{82.7}        & \multicolumn{1}{c|}{59.1}        & \multicolumn{1}{c|}{71.1}         & \multicolumn{1}{c|}{81.7}        & \multicolumn{1}{c|}{60.5}        & \multicolumn{1}{c|}{70.6}         & \multicolumn{1}{c|}{81.9}        & 59.3        \\ \hline
\textbf{EffNet\_b7}             & \multicolumn{1}{c|}{71.1}         & \multicolumn{1}{c|}{81.9}        & \multicolumn{1}{c|}{60.7}        & \multicolumn{1}{c|}{70.5}         & \multicolumn{1}{c|}{82.0}        & \multicolumn{1}{c|}{58.9}        & \multicolumn{1}{c|}{70.7}         & \multicolumn{1}{c|}{80.9}        & \multicolumn{1}{c|}{60.5}        & \multicolumn{1}{c|}{70.2}         & \multicolumn{1}{c|}{81.1}        & 59.4        \\ \hline
\textbf{EffNet\_v2\_s}          & \multicolumn{1}{c|}{71.3}         & \multicolumn{1}{c|}{82.3}        & \multicolumn{1}{c|}{60.4}        & \multicolumn{1}{c|}{70.9}         & \multicolumn{1}{c|}{82.8}        & \multicolumn{1}{c|}{59.0}        & \multicolumn{1}{c|}{70.9}         & \multicolumn{1}{c|}{81.6}        & \multicolumn{1}{c|}{60.2}        & \multicolumn{1}{c|}{70.4}         & \multicolumn{1}{c|}{81.8}        & 58.9        \\ \hline
\textbf{EffNet\_v2\_m}          & \multicolumn{1}{c|}{70.7}         & \multicolumn{1}{c|}{82.2}        & \multicolumn{1}{c|}{60.2}        & \multicolumn{1}{c|}{70.0}         & \multicolumn{1}{c|}{81.6}        & \multicolumn{1}{c|}{58.4}        & \multicolumn{1}{c|}{70.2}         & \multicolumn{1}{c|}{80.5}        & \multicolumn{1}{c|}{60.0}        & \multicolumn{1}{c|}{69.4}         & \multicolumn{1}{c|}{80.6}        & 58.3        \\ \hline
\textbf{EffNet\_v2\_l}          & \multicolumn{1}{c|}{71.4}         & \multicolumn{1}{c|}{82.5}        & \multicolumn{1}{c|}{60.5}        & \multicolumn{1}{c|}{71.0}         & \multicolumn{1}{c|}{82.7}        & \multicolumn{1}{c|}{59.3}        & \multicolumn{1}{c|}{70.8}         & \multicolumn{1}{c|}{81.6}        & \multicolumn{1}{c|}{60.1}        & \multicolumn{1}{c|}{70.2}         & \multicolumn{1}{c|}{82.0}        & 58.4        \\ \hline  \hline
\textbf{Swin\_t}                & \multicolumn{1}{c|}{71.1}         & \multicolumn{1}{c|}{82.0}        & \multicolumn{1}{c|}{60.3}        & \multicolumn{1}{c|}{70.6}         & \multicolumn{1}{c|}{82.3}        & \multicolumn{1}{c|}{58.9}        & \multicolumn{1}{c|}{70.6}         & \multicolumn{1}{c|}{81.3}        & \multicolumn{1}{c|}{59.9}        & \multicolumn{1}{c|}{70.2}         & \multicolumn{1}{c|}{81.6}        & 58.7        \\ \hline
\textbf{Swin\_s}                & \multicolumn{1}{c|}{71.2}         & \multicolumn{1}{c|}{82.4}        & \multicolumn{1}{c|}{60.0}        & \multicolumn{1}{c|}{70.8}         & \multicolumn{1}{c|}{82.7}        & \multicolumn{1}{c|}{58.8}        & \multicolumn{1}{c|}{70.7}         & \multicolumn{1}{c|}{81.7}        & \multicolumn{1}{c|}{59.7}        & \multicolumn{1}{c|}{70.2}         & \multicolumn{1}{c|}{82.0}        & 58.5        \\ \hline
\textbf{Swin\_b}                & \multicolumn{1}{c|}{71.2}         & \multicolumn{1}{c|}{82.5}        & \multicolumn{1}{c|}{60.6}        & \multicolumn{1}{c|}{70.5}         & \multicolumn{1}{c|}{82.8}        & \multicolumn{1}{c|}{58.3}        & \multicolumn{1}{c|}{70.7}         & \multicolumn{1}{c|}{81.8}        & \multicolumn{1}{c|}{59.6}        & \multicolumn{1}{c|}{70.4}         & \multicolumn{1}{c|}{82.0}        & 58.8        \\ \hline
\textbf{Swin\_v2\_t}            & \multicolumn{1}{c|}{71.4}         & \multicolumn{1}{c|}{82.2}        & \multicolumn{1}{c|}{60.6}        & \multicolumn{1}{c|}{70.9}         & \multicolumn{1}{c|}{82.5}        & \multicolumn{1}{c|}{59.3}        & \multicolumn{1}{c|}{71.0}         & \multicolumn{1}{c|}{81.6}        & \multicolumn{1}{c|}{60.4}        & \multicolumn{1}{c|}{70.3}         & \multicolumn{1}{c|}{81.9}        & 58.7        \\ \hline
\textbf{Swin\_v2\_s}            & \multicolumn{1}{c|}{71.1}         & \multicolumn{1}{c|}{82.3}        & \multicolumn{1}{c|}{60.5}        & \multicolumn{1}{c|}{70.6}         & \multicolumn{1}{c|}{81.9}        & \multicolumn{1}{c|}{59.2}        & \multicolumn{1}{c|}{70.7}         & \multicolumn{1}{c|}{81.1}        & \multicolumn{1}{c|}{60.2}        & \multicolumn{1}{c|}{69.8}         & \multicolumn{1}{c|}{80.9}        & 58.8        \\ \hline
\textbf{Swin\_v2\_b}            & \multicolumn{1}{c|}{71.8}         & \multicolumn{1}{c|}{82.5}        & \multicolumn{1}{c|}{61.1}        & \multicolumn{1}{c|}{71.3}         & \multicolumn{1}{c|}{82.8}        & \multicolumn{1}{c|}{59.7}        & \multicolumn{1}{c|}{71.4}         & \multicolumn{1}{c|}{81.9}        & \multicolumn{1}{c|}{60.8}        & \multicolumn{1}{c|}{70.7}         & \multicolumn{1}{c|}{82.0}        & 59.5        \\ \hline  \hline
\textbf{ViT\_b\_16}             & \multicolumn{1}{c|}{70.8}         & \multicolumn{1}{c|}{81.9}        & \multicolumn{1}{c|}{59.7}        & \multicolumn{1}{c|}{70.1}         & \multicolumn{1}{c|}{82.1}        & \multicolumn{1}{c|}{58.2}        & \multicolumn{1}{c|}{70.3}         & \multicolumn{1}{c|}{81.2}        & \multicolumn{1}{c|}{59.5}        & \multicolumn{1}{c|}{69.4}         & \multicolumn{1}{c|}{81.1}        & 57.7        \\ \hline
\textbf{ViT\_b\_32}             & \multicolumn{1}{c|}{67.4}         & \multicolumn{1}{c|}{78.8}        & \multicolumn{1}{c|}{56.4}        & \multicolumn{1}{c|}{66.6}         & \multicolumn{1}{c|}{78.7}        & \multicolumn{1}{c|}{54.6}        & \multicolumn{1}{c|}{66.8}         & \multicolumn{1}{c|}{77.6}        & \multicolumn{1}{c|}{56.0}        & \multicolumn{1}{c|}{66.2}         & \multicolumn{1}{c|}{77.5}        & 54.9        \\ \hline
\textbf{ViT\_l\_32}             & \multicolumn{1}{c|}{69.4}         & \multicolumn{1}{c|}{81.0}        & \multicolumn{1}{c|}{57.8}        & \multicolumn{1}{c|}{68.8}         & \multicolumn{1}{c|}{81.3}        & \multicolumn{1}{c|}{56.2}        & \multicolumn{1}{c|}{68.9}         & \multicolumn{1}{c|}{80.2}        & \multicolumn{1}{c|}{57.6}        & \multicolumn{1}{c|}{68.3}         & \multicolumn{1}{c|}{80.4}        & 56.2        \\ \hline  \hline
\textbf{FUXI }                   & \multicolumn{1}{c|}{74.0}         & \multicolumn{1}{c|}{84.2}        & \multicolumn{1}{c|}{63.8}        & \multicolumn{1}{c|}{73.5}         & \multicolumn{1}{c|}{84.5}        & \multicolumn{1}{c|}{62.4}        & \multicolumn{1}{c|}{73.6}         & \multicolumn{1}{c|}{83.6}        & \multicolumn{1}{c|}{63.5}        & \multicolumn{1}{c|}{73.2}         & \multicolumn{1}{c|}{84.0}        & 62.4        \\ \hline
\textbf{SITU }                   & \multicolumn{1}{c|}{71.1}         & \multicolumn{1}{c|}{82.0}        & \multicolumn{1}{c|}{60.3}        & \multicolumn{1}{c|}{70.5}         & \multicolumn{1}{c|}{82.2}        & \multicolumn{1}{c|}{58.8}        & \multicolumn{1}{c|}{70.6}         & \multicolumn{1}{c|}{81.2}        & \multicolumn{1}{c|}{59.9}        & \multicolumn{1}{c|}{70.0}         & \multicolumn{1}{c|}{81.2}        & 58.8        \\ \hline
\textbf{CTC }                    & \multicolumn{1}{c|}{71.0}         & \multicolumn{1}{c|}{82.2}        & \multicolumn{1}{c|}{59.8}        & \multicolumn{1}{c|}{70.3}         & \multicolumn{1}{c|}{82.4}        & \multicolumn{1}{c|}{58.2}        & \multicolumn{1}{c|}{70.6}         & \multicolumn{1}{c|}{81.5}        & \multicolumn{1}{c|}{59.6}        & \multicolumn{1}{c|}{70.1}         & \multicolumn{1}{c|}{81.8}        & 58.5        \\ \hline
\end{tabular}}
\end{table}

Table \ref{table7} presents the performance on the new AffectNet-AV partition, showing consistent rankings across all models for Global CCC$^{\text{VA}}$, CCC$^{\text{V}}$, and CCC$^{\text{A}}$. Among the baseline models, ConvNeXt\_b achieves the highest performance, followed by EffNet\_b6, VGG19, iResNet101, ResNext101\_64, DenseNet161, and ResNet34. Within transformer models, the Swin Transformer outperforms the ViT variant, consistent with other tasks. Among the ABAW methods, FUXI outperforms SITU, which is followed by CTC. This ranking differs from previous results and reinforces our conclusions drawn from the other two tasks. Additionally, For the fairness evaluation, the Local CCC on valence, arousal, and valence-arousal aligns with the model performance rankings of the Global CCC$^{\text{VA}}$. The Local CCC$^{\text{VA}}$ for gender and race is closest to the Global CCC$^{\text{VA}}$, with an average difference of around 0.5\%. However, for age, the difference exceeds 1\%, indicating greater bias. Regarding Local CCC$^{\text{A}}$, the difference for gender is within 0.5\%, while for race and age, it exceeds 1.5\%. Interestingly, for Local CCC$^{\text{V}}$, the score for race is higher than the Global score across all models. For example, FUXI's Local score is 84.5, while its Global score is 84.2. This discrepancy can be attributed to data imbalance, causing the model to perform better on more represented race groups, leading to extreme performances within certain race groups and affecting the overall Global CCC$^{\text{VA}}$.




\vspace{-10pt}

\section{CONCLUSION}
In conclusion, this research highlights the importance of addressing biases and promoting fairness in automatic affect analysis. The study analyzed six affective databases, annotated demographic attributes, and proposed a protocol for fair database partitioning. Extensive experiments with baseline and SOTA methods revealed the significant impact of these changes and the inadequacy of prior assessments. Emphasizing demographic attributes and fairness in affect analysis research provides a foundation for enhancing model performance and developing more equitable methodologies.

\ifCLASSOPTIONcaptionsoff
  \newpage
\fi

\bibliographystyle{IEEEtran}
\bibliography{egbib}

\end{document}